%% file: eccv_main.tex
\documentclass[runningheads]{llncs}

\usepackage{eccv}

\usepackage{graphicx}
\usepackage{caption}
\usepackage{subcaption}
\usepackage{adjustbox}
\usepackage{makecell}
\usepackage{booktabs}
\usepackage{xspace}
\usepackage{pifont}
\usepackage{colortbl}
\usepackage{floatrow}
\usepackage{xcolor}
\usepackage[bottom]{footmisc}

\usepackage{lipsum}
\usepackage{multicol}
\usepackage{xspace}
\usepackage{ulem}
\usepackage{tikz,pgfplots,tkz-kiviat}
\usepackage{arrayjobx}
\usetikzlibrary{decorations.pathreplacing, arrows, fit,backgrounds,positioning,shapes,plotmarks,calc,decorations,angles,decorations.markings,intersections}
\usepackage{multirow}
\usepackage{mathtools}
\usepackage{amssymb}
\usepackage{amsmath}
\usepackage{amsfonts}
\usepackage{bm}
\usepackage{subcaption}
\usepackage{booktabs}
\usepackage{wrapfig}
\usepackage[accsupp]{axessibility}

\usepackage{float}
\floatstyle{plaintop}
\restylefloat{table}
\usepackage{hyperref}

\usepackage{orcidlink}

\input{abbrev}

\begin{document}

\title{\ours: Multi-Person Whole-Body \\Human Mesh Recovery in a Single Shot}

\titlerunning{\Ours}

\author{Fabien Baradel$^{*}$ \and Matthieu Armando \and Salma Galaaoui \and Romain Br\'egier \and \\Philippe Weinzaepfel \and Gr\'egory Rogez \and Thomas Lucas$^{*}$}

\authorrunning{Baradel et al.}

\institute{
NAVER LABS Europe \\[0.01cm]
\url{https://github.com/naver/multi-hmr} \\[0.01cm] $^{*}$Equal contribution
}

\maketitle

\input{sections/00_abstract}
\input{sections/01_intro}
\input{sections/02_related}
\input{sections/03_method}

\input{sections/04_expe}

\input{sections/05_conclusion}

\bibliographystyle{splncs04}
\bibliography{eccv_main}

\clearpage
\appendix

\input{sections/99_suppmat}

\end{document}

%% file: sections/00_abstract.tex
\begin{abstract}
We present \Ours, a strong \onestage model for \emph{multi-person} 3D human mesh recovery from a single RGB image.
Predictions encompass the \emph{whole body}, \ie, including hands and facial expressions, using the SMPL-X parametric model and
\emph{3D location} in the camera coordinate system.
Our model detects people by predicting coarse 2D heatmaps of person locations, using features produced by a standard Vision Transformer (ViT) backbone. 
It then predicts their whole-body pose, shape and 3D location using a new cross-attention module called the Human Prediction Head (HPH), with one query attending to the entire set of features for each detected person.
As direct prediction of fine-grained hands and facial poses in a single shot, \ie, without relying on explicit crops around body parts, is hard to learn from existing data, we introduce \dataset, the \datasetname dataset, containing humans close to the camera with diverse hand poses. We show that incorporating it into the training data further enhances predictions, particularly for hands.
\Ours also optionally accounts for \emph{camera intrinsics}, if available, by encoding camera ray directions for each image token.
This simple design achieves strong performance on whole-body and body-only benchmarks simultaneously: a ViT-S backbone on $448{\times}448$ images already yields a fast and competitive model, while 
larger models and higher resolutions obtain state-of-the-art results.

\begin{figure}[b]
\centering
\begin{tabular}{cc}
\includegraphics[width=0.65\textwidth,trim={0 1.5cm 0 0},clip]{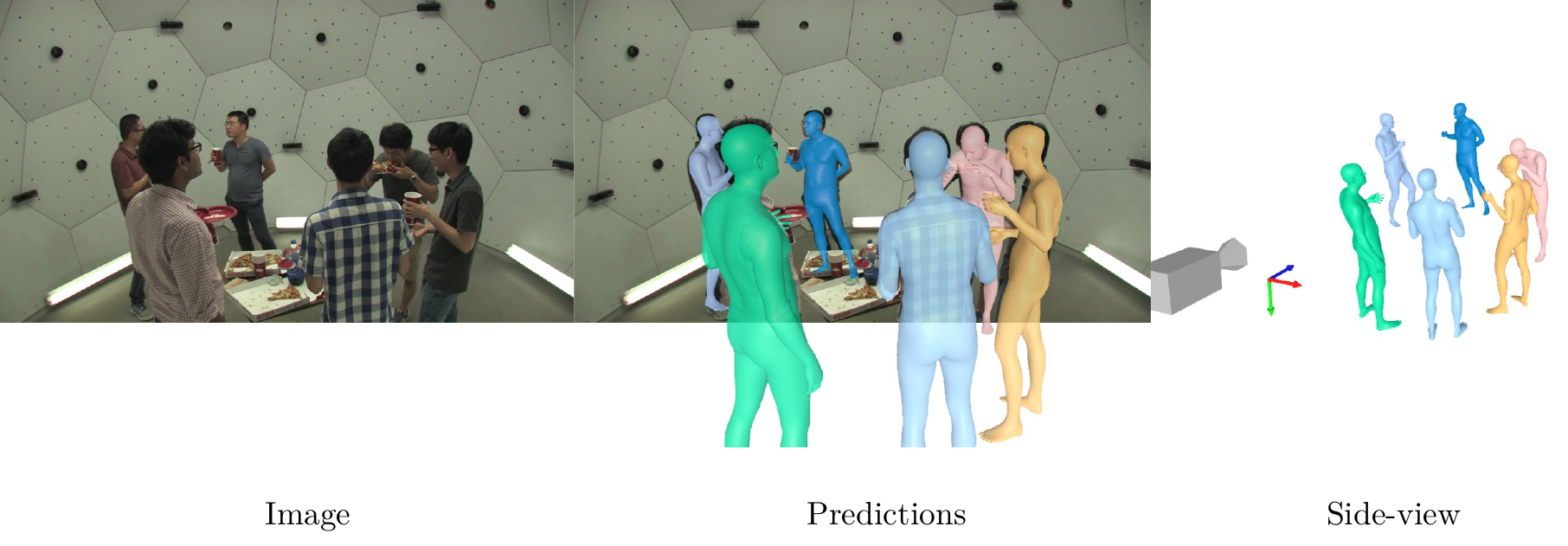} &
\adjustbox{width=0.34\linewidth}{
\input{tab/backbone_spider}
} \\[-0.1cm]
\small \hspace{1cm} Image \hspace{1.6cm} Predictions \hspace{0.8cm} Side View & \small Relative improvement (\%) 
\end{tabular}
\\[-0.3cm]
\caption{
 \textbf{Efficient 3D reconstruction of multiple humans in camera space.}
We introduce \Ours, a \onestage approach to detect \textit{multiple humans}
in images, and regress \textit{whole-body} human meshes. Predictions encompass hands and facial expressions, as well as 3D location with respect to the camera. 
\textit{Left:} Visualization of \Ours predictions. \textit{Right:} Relative improvements (in \%) on human mesh recovery benchmarks.
}
\label{fig:teaser}
\end{figure}

\end{abstract}

%% file: tab/backbone_spider.tex
\newarray\kivaxisitemlabels
\readarray{kivaxisitemlabels}{%
SKIP &    &      &       &  &
\Huge 3DPW (PVE)    &
     &      &   &  &
  \Huge   MuPoTs (PCK3D)    &
     &      &   &  &
\Huge  CMU (MPJPE) &
    &  &  &  &
\Huge EHF (PVE)   &
    &      &   &  &
\Huge  AGORA-X (MVE)  &
     &      &   &  &
\Huge UBody (PVE)  &
0\%    & 20\%    &  40\% & XX &
\Huge AGORA (PVE)  &
}

\dataheight=5

\newcommand{\kivcurrentlabel}[2]{\checkkivaxisitemlabels(#1,#2)\trimspace\cachedata \cachedata}

\newcommand{\kivaxisnumbers}{7}
\newcommand{\kivcategorycounts}{{3,3,3,3,3,3,3}}
\newcommand{\kivlcmcatcount}{3}

\newcommand{\kivlattice}{%
\pgfmathsetmacro{\kivaxisangle}{360/\kivaxisnumbers}
    \foreach \x in {1,...,\kivaxisnumbers}
    {   \foreach \y in {1,...,\kivlcmcatcount}
        {   \pgfmathsetmacro{\kivaxisstep}{6/\kivlcmcatcount}
            \draw[help lines,gray,line width=0.1mm] (\kivaxisangle*\x:\y*\kivaxisstep) -- (\kivaxisangle*\x+\kivaxisangle:\y*\kivaxisstep);
        }
    }
    \foreach \x in {1,...,\kivaxisnumbers}
    {   \draw[-,gray,line width=0.1mm] (0,0) -- (\kivaxisangle*\x:6cm);
        \pgfmathsetmacro{\kivaxissteps}{\kivcategorycounts[\x-1]}
        \pgfmathsetmacro{\kivaxisstep}{6/\kivcategorycounts[\x-1]}
        \pgfmathtruncatemacro{\kivlabelnumber}{1+1}  
        \pgfmathtruncatemacro{\ly}{1} 
         \pgfmathtruncatemacro{\morespace}{1} 
            \node[label=\x*\kivaxisangle:{\Huge \kivcurrentlabel{\x}
            {\kivlabelnumber}},inner sep=1pt] at (\kivaxisangle-11*\x:\kivaxisstep*\morespace) {};
        \pgfmathtruncatemacro{\kivlabelnumber}{2+1}  
        \pgfmathtruncatemacro{\ly}{2} 
        \pgfmathtruncatemacro{\morespace}{\ly+0.3} 
        \node[label=\x*\kivaxisangle:{\Huge \kivcurrentlabel{\x}
            {\kivlabelnumber}},inner sep=1pt] at (\kivaxisangle-11*\x:\kivaxisstep*\ly) {};
        \foreach \y in {3,...,\kivaxissteps}
        {
            \pgfmathtruncatemacro{\kivlabelnumber}{\y+1}  
            \node[label=\x*\kivaxisangle:{\Huge \kivcurrentlabel{\x}
            {\kivlabelnumber}},inner sep=1pt] at (\kivaxisangle-11*\x:\kivaxisstep*\y) {};  
        }
        \pgfmathtruncatemacro{\lasty}{\kivaxissteps+3}
        \pgfmathsetmacro{\lastyplace}{\kivaxissteps+0.1}
        \pgfmathtruncatemacro{\kivlabelnumber}{\lasty}  
            \node[label=\x*\kivaxisangle:{\Huge \kivcurrentlabel{\x}{\kivlabelnumber}},inner sep=1pt] at (\kivaxisangle*\x:\kivaxisstep*\lastyplace) {};
    }
}

\newcommand{\kivdatapoints}{}

\newcommand{\kivdata}[2]{
    \renewcommand{\kivdatapoints}{{#1}}
    \pgfmathsetmacro{\kivaxisangle}{360/\kivaxisnumbers}
    \pgfmathsetmacro{\kivcoordinate}{\kivdatapoints[0]*6}
    \fill[opacity=0.1,#2] (\kivaxisangle:\kivcoordinate) 
        \foreach \x in {1,...,\kivaxisnumbers}
        {   
            -- (\kivaxisangle*\x:\kivdatapoints[\x-1]*6) 
        }
    -- cycle;
    \draw[#2,line width=1.5mm] (\kivaxisangle:\kivcoordinate) 
        \foreach \x in {1,...,\kivaxisnumbers}
        {  
            -- (\kivaxisangle*\x:\kivdatapoints[\x-1]*6) 
        }
    -- cycle;
}

\begin{tikzpicture}
\kivlattice
\kivdata{0.505, 0.609, 0.773, 0.949, 0.511, 0.844, 0.619}{NavyBlue}
\kivdata{0.33,0.33,0.33,0.33,0.33,0.33, 0.33}{Bittersweet}
\node[draw, Bittersweet, minimum width=1cm, minimum height=0.07cm, fill, label=right:{\Huge Prev. SotA}] at (6,3) {};
\node[draw, NavyBlue, minimum width=1cm, minimum height=0.07cm, fill, label=right:{\Huge \Ours}] at (6,4) {};
\draw (5,2.3) rectangle (11.5,4.7);
\end{tikzpicture}

%% file: sections/01_intro.tex
\section{Introduction}

We introduce a \onestage model for recovering whole-body 3D meshes of humans from a single RGB image. 
Our problem formulation focuses on four aspects of Human Mesh Recovery (HMR) that we identify as pivotal to making HMR applicable to real-world scenarios:
i) capture of expressive body poses -- \ie, including hands and facial expressions,
ii) efficient processing 
of images with a variable number of people,
iii) 
location of people in 3D space,
iv) adaptability to camera information 
when
available.

Successfully handling these aspects 
simultaneously makes 
our proposed model, denoted \Ours, widely applicable.
For instance, in virtual or augmented reality (AR/VR), capturing faces and hands precisely 
is key for communication.
It is also beneficial for enabling human-robot interactions~\cite{salzmann2023robots,de2008atlas},
or human understanding from images and videos~\cite{rajasegaran2023benefits,zhou2023human,shah2022pose}.
Likewise, understanding the placement of people in the scene is necessary for applications ranging from robotic navigation to AR/VR applications involving several people. 
In addition, efficient processing of a variable number of people is desirable when computation is restricted or real-time processing is needed. 
Finally, reasoning about 3D meshes can only benefit from adapting to camera information when it is available~\cite{kocabas2021spec,li2022cliff}.   

\def\faChecked{\FA\symbol{"F00C}}
\def\faCrossed{\FA\symbol{"F00D}}

\newcommand{\colcell}[1]{\cellcolor{green!10}{#1}}
\setlength{\intextsep}{5pt}
\begin{wraptable}{r}{6.3cm}
\caption{
\textbf{Main features} of \Ours \textit{vs.} the state of the art:
    Single-person methods rely on human detectors to process image crops around each person independently.
    Multi-person approaches detect humans and regress their properties using the same network.
    \emph{\Onestage} refers to methods regressing the expected output without extracting or resampling features from different regions.
    \\[-0.75cm]
    }
    \label{tab:feature}
\setlength{\tabcolsep}{2pt}
\renewcommand{\arraystretch}{1.1}
\resizebox{1\linewidth}{!}{\input{tab/03_features}}
\end{wraptable}

In their pioneering work on HMR~\cite{kanazawa2018hmr}, Kanazawa \etal propose to predict SMPL mesh parameters and three parameters for weak-perspective reprojection given a cropped image containing a person. 
Different aspects of this approach have been improved since, including architectures ~\cite{goel2023humans4d,li2022cliff,zhang2021pymaf},  training techniques~\cite{kolotouros2019spin} and data enhancements~\cite{agora,bedlam,joo2020eft}.
The approach has also been extended to whole-body parametric models like SMPL-X~\cite{ehf}, often with multiple crops centered on body, hands and face~\cite{choutas2020expose,pixie,hand4whole}.
Multi-person inputs are typically handled with a two-step procedure: first running an off-the-shelf human detector, then applying a mesh recovery model on crops around each detected person.
Conversely, ROMP~\cite{romp} and PSVT~\cite{psvt} recover multiple human meshes in a single step using one-shot detectors. BEV~\cite{bev} additionally predicts the relative depths of meshes.
Accounting for intrinsic camera parameters has been shown to  improve reprojection~\cite{kocabas2021spec,li2022cliff}, especially when these differ between training and inference.
Despite these advancements, no previous method has successfully integrated in a single model all four essential features: 
efficient multi-person processing, whole-body mesh recovery, location estimation in camera space and, optionally, camera-aware predictions. 
Please refer to Table~\ref{tab:feature} for a comparison to existing work.

\begin{figure*}[t]
\includegraphics[width=\linewidth]{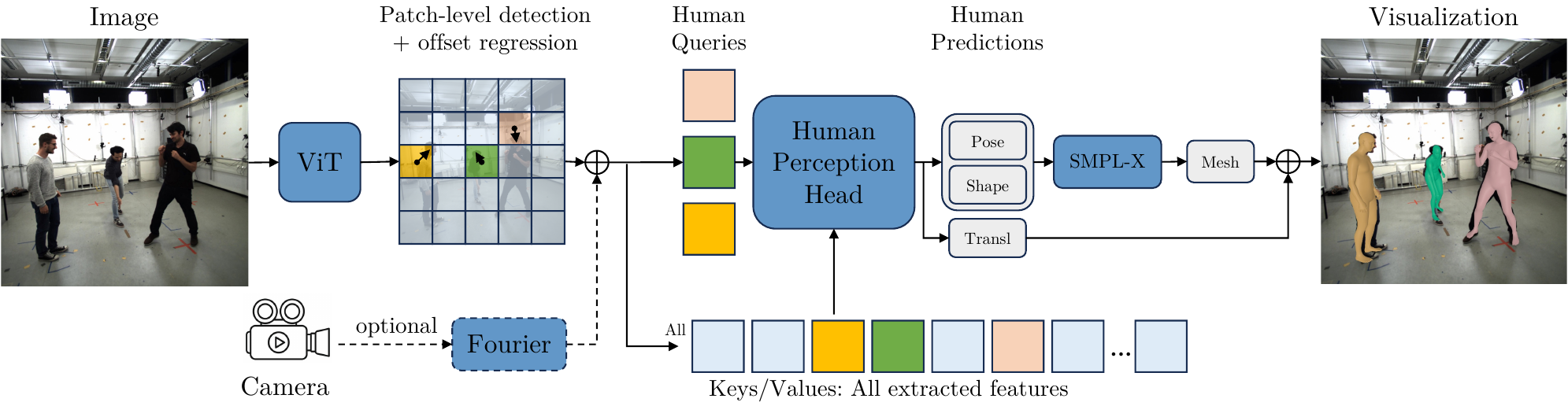} \\[-0.35cm]
\caption{\textbf{Overview of \Ours.} 
A ViT backbone extracts image embeddings.
Detection is conducted at the patch level with additional 2D offset regression. Each detected token serves as a query for a cross-attention-based head, called the Human Perception Head (HPH), which predicts pose and shape parameters, along with location in 3D space. Optionally, known camera parameters are embedded and added to each patch, represented as a Fourier encoding of the ray originating from the camera center.
}
\label{fig:archi}
\end{figure*}

In this paper, we introduce \Ours, an efficient \onestage method that detects each person in a scene and regresses their pose, shape, and 3D location in camera space, using a whole-body parametric mesh model.
Please see Figure~\ref{fig:teaser} (left) for an example of prediction.
Optionally, \Ours can be conditioned on camera intrinsics if available.
Figure~\ref{fig:archi} presents an overview of the model architecture.
We use a standard Vision Transformer (ViT)~\cite{dosovitskiy2020image} backbone to extract features from the input data, which allows us to benefit from recent advancements in large-scale self-supervised pre-training~\cite{MaskedAutoencoders2021,dinov2,dinov1}.
This differs from architectures like HR-Net~\cite{hrnet} which are less common in the pre-training literature.
We regress a person-center heatmap from the feature tensor produced by the 
backbone:
for each input token, the model first outputs a probability that a person is centered on a point present in the corresponding input patch, as well as location offsets~\cite{zhou2019objects}.
We introduce a prediction head called the Human Perception Head (HPH) that employs cross-attention.  In this mechanism, queries correspond to the detected center tokens, while keys and values are drawn from all image tokens. 
It efficiently predicts pose and shape parameters of an expressive human model, namely SMPL-X~\cite{choutas2020expose}, for a variable number of detections,
while also regressing depths to place individuals within the scene.
To improve 3D prediction by incorporating camera intrinsics, our model can optionally take camera parameters as input. These parameters are used to augment each token feature with Fourier-encoding of the corresponding camera ray directions before passing them to the prediction head.

\Ours is conceptually simple: unlike 
most existing whole-body approaches, it does not rely on multiple high-resolution crops of the body parts for expressive models~\cite{pixie,hand4whole,choutas2020expose}, or hand-designed components to place people in the scene~\cite{romp,3dcrowdnet}.
However, naively regressing SMPL-X parameters from a single token feature tends to 
under-perform
on small body parts like hands.
We find that incorporating expressive human subjects positioned close to the camera in the training data results in good performance across all body parts. 
We thus introduce the \dataset (\datasetname) dataset, containing synthetic renderings of people with clearly visible hands in diverse poses.

We train 
a family of models
with various backbone sizes and input resolutions.
We evaluate performance on 
both body-only (3DPW~\cite{3dpw}, MuPoTs~\cite{mupots}, CMU-Panoptic~\cite{cmu}, AGORA-SMPL~\cite{agora}) and whole-body expressive mesh recovery benchmarks (EHF~\cite{ehf}, AGORA-SMPLX~\cite{agora} and UBody~\cite{osx}), see Figure~\ref{fig:teaser} (right).
The \onestage nature of the model allows for efficient inference. For instance, with a ViT-S backbone and $448{\times}448$ inputs, \Ours 
is competitive on both body-only and whole-body datasets while being real-time, achieving $30$ frames per second (fps) on a NVIDIA V100 GPU.
Larger backbones and higher resolutions -- up to a ViT-L backbone and $896{\times}896$ inputs -- outperform the state of the art at the cost of slower but still reasonable inference speed (5 fps).

%% file: tab/03_features.tex
\begin{tabular}{r@{}l@{~~}c@{~~}c@{~~}c@{~~}c}
    \cmidrule[\heavyrulewidth](lr){1-6}
   &  \multirow{2}{*}{Method}    & Whole & Single & Camera & Camera \\
  &       & Body & Shot & Space & Aware  \\
         \cmidrule(lr){1-6}
  \multirow{9}{*}{$\begin{aligned} \text{Single-}\text{person} \end{aligned} \begin{dcases*} \\ \\ \\ \\ \\ \\ \\ \\ \end{dcases*}$\hspace*{0.15cm}}
                                                                                                                                & HMR~\cite{kanazawa2018hmr}                      & \redxmark   & \greencmark & \redxmark   & \redxmark                          \\
                                                                                                                                & HMR2.0~\cite{goel2023humans4d}                  & \redxmark   & \greencmark & \redxmark   & \redxmark                          \\
                                                                                                                                & SPEC~\cite{kocabas2021spec}                     & \redxmark   & \greencmark & \redxmark   & \greencmark                        \\
                                                                                                                                & CLIFF~\cite{li2022cliff}                        & \redxmark   & \greencmark & \redxmark   & \greencmark                        \\
                                                                                                                                & PIXIE~\cite{pixie}                              & \greencmark & \redxmark   & \redxmark   & \redxmark                        \\
                                                                                                                                & Hand4Whole~\cite{hand4whole}                    & \greencmark & \redxmark   & \redxmark   & \redxmark                        \\
                                                                                                                                & PyMAF-X~\cite{pymafx2023}                       & \greencmark & \redxmark   & \redxmark   & \redxmark                        \\
                                                                                                                                & OSX~\cite{osx}                                  & \greencmark & \redxmark   & \redxmark   & \redxmark                        \\
                                                                                                                                & SMPLer-X~\cite{cai2023smplerx}                  & \greencmark & \redxmark   & \redxmark   & \redxmark                        \\
 \raisebox{-0.6ex}{Det. + Single} { \raisebox{-.7ex}{\Large \{\kern.4ex}} \hspace*{0.04cm}                                       & \raisebox{-.25ex}{3DCrowdNet~\cite{3dcrowdnet}} & \redxmark   & \redxmark   & \redxmark   & \redxmark                        \\
 \multirow{4}{*}{\textcolor{ForestGreen}{Multi-person}  $\begin{dcases*} \\ \\ \\ \end{dcases*}$\hspace*{0.15cm}}                                        & \colcell{ROMP~\cite{romp}}                                & \redxmark   & \greencmark & \redxmark   & \redxmark                          \\
                                                                                                                                & \colcell{BEV~\cite{bev}}                                  & \redxmark   & \greencmark & \greencmark & \redxmark                         \\
                                                                                                                                & \colcell{PSVT~\cite{psvt}}                                & \redxmark   & \greencmark & \greencmark & \redxmark                         \\
                                                                                                                                & \colcell{\bf{\Ours}}                                      & \greencmark & \greencmark & \greencmark & \textcolor{ForestGreen}{Optional}  \\
        \cmidrule[\heavyrulewidth](lr){1-6} \addlinespace[-\belowrulesep] 
    \end{tabular}

%% file: sections/02_related.tex
\section{Related work}
\label{sec:related}

\Ours primarily builds upon whole-body HMR and multi-person HMR. It also relies on synthetic datasets.
We now review these three literatures. 

\myparagraph{Whole-body Human Mesh Recovery}
There has been a recent surge of interest
for whole-body mesh recovery from a single image~\cite{ehf,hand4whole,osx,pixie}, 
fostered in part by seminal work on improving whole-body parametric models.
In particular SMPL-X~\cite{ehf}
outputs an expressive mesh for the whole body given a small set of pose and shape parameters.
The first approaches 
were based on optimization, \eg SMPLify-X~\cite{ehf}, but they remain slow and sensitive to local minima.
Numerous learning-based methods were also introduced,
but only in single-person settings~\cite{hand4whole,pixie,choutas2020expose,rong2021frankmocap,zhou2021monocular,pymafx2023,cai2023smplerx}.
This setting already poses significant challenges: hands and faces are typically low resolution in natural images, and capturing their poses hinges on subtle details.
To overcome this, most approaches leverage a multi-crop pipeline: areas of interest -- such as the face and hands -- are cropped, resized and used to estimate the associated meshes which are aggregated into a whole-body prediction.
In particular, ExPose~\cite{choutas2020expose} selects high-resolution crops using a body-driven attention mechanism;
PIXIE~\cite{pixie} fuses body parts in an adaptive manner, and
Hand4Whole~\cite{hand4whole} uses both body and hand joint features for 
3D wrist rotation estimation. 
In contrast to these methods, \Ours is \onestage, without high-resolution crops. 
More recently, OSX~\cite{osx} 
 introduced the first single-crop method for single-person whole-body mesh recovery.
They leverage a ViT encoder, followed by a high-resolution feature pyramid, and use keypoint (\eg wrists) estimates to resample features in their decoder head. 
SMPLer-X~\cite{cai2023smplerx} employed a similar approach, training on numerous datasets.
We depart from existing methods by i) tackling \emph{multi-person} whole-body mesh recovery
and ii) using a \onestage approach, with a non-hierarchical feature extractor.

\myparagraph{Multi-Person Human Mesh Recovery}
Most existing multi-person HMR methods~\cite{3dcrowdnet,goel2023humans4d,kolotouros2019spin,zhang2021pymaf,qiu2022dynamic} build upon a multi-stage framework: an off-the-shell human detector~\cite{redmon2016you,he2017mask,liu2016ssd} is used, followed by a single-person mesh estimation model~\cite{Ma_2023_CVPR,Kim_2023_CVPR,Yoshiyasu_2023_CVPR,Zheng_2023_CVPR} to process each detected human.
This has two drawbacks: i) it is inefficient at inference time compared to a \onestage approach and ii) the pipeline cannot be 
learned
end-to-end. This impacts final performance, in particular in cases of truncation by the image frame or person-person occlusions, a common scenario in multi-person settings. 
Following the seminal work of ROMP~\cite{romp} which estimates 2D maps for 2D human detections, positions and mesh parameters, single-stage models have been proposed~\cite{romp,bev,psvt}.
In particular, BEV~\cite{bev} introduces an additional Bird-Eye-View representation of the scene to predict relative depth between detected persons and
PSVT~\cite{psvt} improves performance using a transformer decoder.
We follow the same single-shot philosophy as~\cite{romp,bev,psvt} but go beyond their settings by: i) tackling whole-body mesh recovery, ii) regressing the 3D location  of each person in the camera coordinate system, and iii) incorporating camera intrinsics as an optional input.
We also introduce an efficient cross-attention-based head, making \Ours faster to train, efficient at inference and improving performance.

\myparagraph{Synthetic data}
Acquiring high-quality real-world ground-truth data at scale for human mesh recovery is costly, in particular when considering faces and hands expressions.
A body of work~\cite{varol17_surreal,yang2023synbody,hasson19_obman} has explored the generation of large-scale synthetic data for human-related tasks.
In this work, we experiment with BEDLAM~\cite{bedlam} and AGORA~\cite{agora}, and confirm empirically that using large-scale synthetic data is beneficial for whole-body human mesh regression, compared to real-world data with pseudo ground-truth fits.
We also propose  a new synthetic dataset, \dataset,  which stands for \datasetname, designed to improve performance particularly on hands for one-stage whole-body prediction.
It departs from existing ones in that it contains humans with diverse and clearly visible hand poses, seen from a limited distance, to allow fine details to be captured. 
Our experiments show that this type of training data is key to allow regressing whole-body meshes in a single shot.

%% file: sections/03_method.tex
\section{\Ours}
\label{sec:method}

We now describe our \onestage multi-person whole-body human mesh recovery approach.
Given an input RGB image $\rmI \in \mathbb{R}^{H \times W \times 3}$ with resolution $H \times W$, our model, denoted $\mathcal{H}$, directly outputs a set of $N$ centered whole-body 3D humans meshes $\rmM  \in \mathbb{R}^{V \times 3}$ composed of V vertices, along with their corresponding root 3D locations $\rvt \in \mathbb{R}^{3}$ in the camera coordinate system: 
\begin{equation}
    \big\{ \rmM_n + \rvt_n\big\}_{n \in \{1, \hdots , N\}} = \mathcal{H}(\rmI).
\end{equation}

As preliminaries, Section~\ref{sub:prelim} presents the 3D whole-body parametric model and the camera model that we use.
We then detail the model architecture in Section~\ref{sub:archi} and the training losses in Section~\ref{sub:losses}.

\subsection{Preliminaries}
\label{sub:prelim}

\myparagraph{Human whole-body mesh representation} We build upon the SMPL-X parametric 3D body model~\cite{choutas2020expose}.
Given input parameters for the pose $\vtheta \in \mathbb{R}^{53 \times 3}$ (global orientation, body, hands and jaw poses) 
in axis-angle representation, shape $\vbeta \in \mathbb{R}^{10}$ and facial expression $\valpha \in \mathbb{R}^{10}$, it outputs an expressive human-centered 3D mesh $\rmM = \text{SMPL-X}(\vtheta,\vbeta,\valpha) \in \mathbb{R}^{V \times 3}$, with $V=10,475$ vertices.
The mesh $\rmM$ is centered around a \textit{primary} keypoint -- in this work we choose the head as primary keypoint. 
It is placed in the 3D scene by putting the primary keypoint at the 3D location $\rvt=(t_x,t_y,t_z)$.
For simplicity, let $\rvx=[\vtheta,\vbeta,\valpha]$: the problem reduces to predicting $\rvx$ and $\rvt$ for all detected humans.

\myparagraph{Pinhole camera model}
We assume a simple pinhole camera model to project 3D points on the image plane.
Ignoring distortion, it is defined by an intrinsic matrix  $\rmK \in \mathbb{R}^{3 \times 3}$ of focal length $f$ and principal point $(p_u, p_v)$.
We set the camera pose to the origin.
We have: 
\begin{equation}
 \hspace{-0.5em} \rmK = \left[ \begin{smallmatrix}
f & 0 & p_u \\
0 & f & p_v \\
0 & 0 & 1
\end{smallmatrix} \right]
\text{ and }
\begin{cases}
 & \hspace{-1em} [c_u, c_v, 1]^T = (1/t_z) \cdot \rmK \ [t_x,t_y,t_z]^T \\
 & \hspace{-1em} [t_x,t_y,t_z]^T = t_z \cdot \rmK^{-1} \ [c_u, c_v, 1]^T
 \end{cases}
 ,
 \hspace{-0.6em} \label{eq:proj}
\end{equation}
with $\rvc=(c_u,c_v)$ the 2D image coordinates of the projection of a 3D point
$\rvt$ into the image plane. 
$\rmK$ can thus be used to backproject a 2D point into 3D given its depth $t_z$.
We denote by $\pi_\rmK$ the camera projection operator and $\pi_\rmK^{-1}$ 
its inverse.

\subsection{\Onestage architecture}
\label{sub:archi} 
Our method is summarized in Figure~\ref{fig:archi}. We first encode images into token embeddings using a ViT backbone. These embeddings are used to detect humans and can optionally be combined with camera embeddings. Our proposed Human Perception Head is then employed to regress whole-body human meshes and depth for a variable number of detected humans. 

\myparagraph{ViT backbone}
The input RGB image $\rmI$ is encoded with a ViT backbone~\cite{dosovitskiy2020image}.
It is sub-divided into image patches of size $P\times P$, each embedded into tokens with a linear transformation and positional encoding.
The set of tokens is processed with self-attention blocks into $\rmE \in \mathbb{R}^{H/P \times W/P \times D}$ with $D$ the feature dimension.
The ViT model keeps a constant resolution throughout so that each output token spatially corresponds to a patch in the input image.

\myparagraph{Patch-level detection} 
To detect humans in the input image, we define a \emph{primary keypoint} on human bodies, here the 3D keypoint of the \emph{head} as defined according to the SMPL-X body model.
For each patch index $(i, j) \in \{1, \hdots, H/P\} \times \{1, \hdots, W/P\}$, we predict if the patch centered at $\rvu^{i,j} = (u^i, v^j)$ contains a primary keypoint~\cite{zhou2019objects}, with a score $s^{i,j} \in [0,1]$ computed from the associated token embedding $\rmE^{i, j} \in \mathbb{R}^{D}$
using a Multi-Layer-Peceptron (MLP).
At inference, we apply a threshold $\tau$ to the scores to detect patches containing primary keypoints: 
\begin{equation}
    \big\{\rvu_n\big\}_{n} = \big\{\rvu^{i,j} | s^{i,j} \ge \tau \big\}.
    \label{eq:detection}
\end{equation}
At train time, the ground-truth detections are used for the rest of the model.

\myparagraph{Image coordinates regression}
Detecting people at the patch level yields a rough estimation of the 2D location of the primary keypoint, up to the size of the predefined patch size $P$.
We refine the 2D location of the primary keypoint 
by regressing a residual offset $\rvdelta = (\rvdelta_u, \rvdelta_v)$ from the center of a patch $(u^i, v^j)$, using an MLP taking the corresponding token embedding $\rmE^{i, j}$ as input.
The 2D coordinates predicted for the primary keypoint detected at patch location $(i,j)$ are thus given by:
\begin{equation}
\rvc^{i,j} = [ u^i + \rvdelta_u, v^j + \rvdelta_v ].
\label{eq:loc}
\end{equation}

\begin{figure}[t]
\centering
\subfloat[
\textbf{Human Perception Head}
\label{fig:mucross}
]{
\resizebox{0.62\linewidth}{!}{\includegraphics[width=\linewidth]{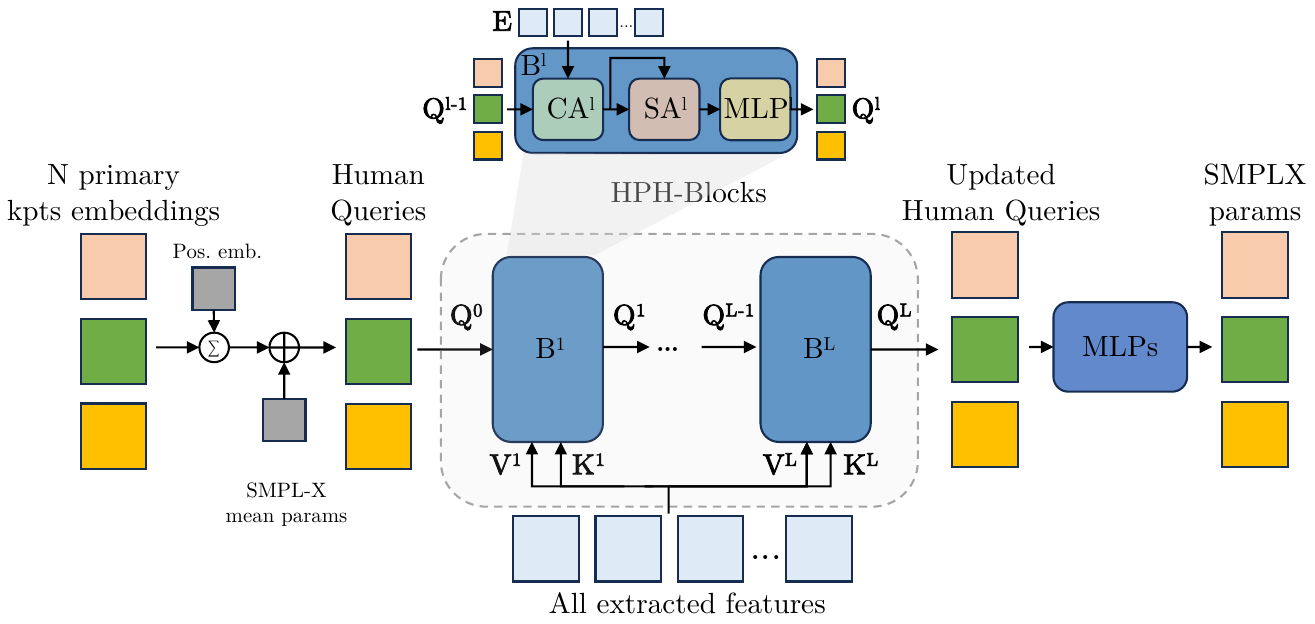}
}}
\subfloat[
\textbf{Samples from \dataset}
\label{fig:cuffs}
]{
\includegraphics[width=0.37\linewidth]{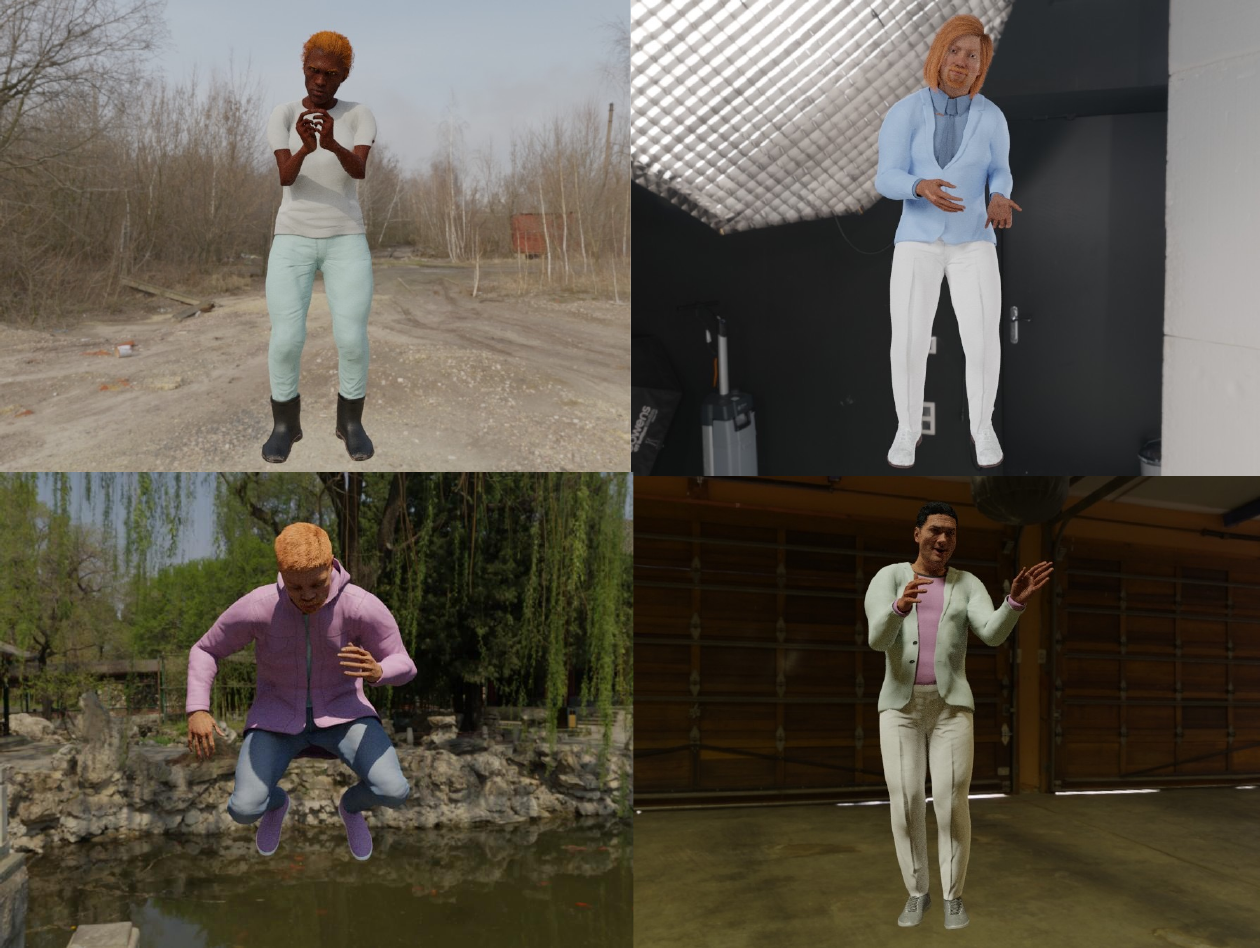}
}
\\[-0.2cm]
\caption{
\textbf{(a)} The token embeddings corresponding to the $N$ detected primary keypoints
are used as queries
in a series of cross-attention blocks where keys and values correspond to the context provided by all image tokens.
MLPs then predict the SMPL-X parameters (pose and shape) as well as the depth for each query. \textbf{(b)} Samples from our CUFFS synthetic dataset.
}
\label{tab:ablations}
\end{figure}

\myparagraph{Human Perception Head (HPH)}
We predict human-centered meshes and depths estimations for all people detected in the scene in a structured manner and in parallel, by processing $\rmE$ with our Human Perception Head, built from cross-attention blocks~\cite{jaegle2021perceiver}, see
Figure~\ref{fig:mucross} for an overview.
This design choice allows features corresponding to a person detection to attend information from all image patches before making a full pose, shape and depth prediction for this person.
For a human detection $n$ at patch location $(i, j)$, we initialize a cross-attention query $\rvq_n = (\rmE^{i,j} \oplus \overline{\rvx}) + \rvp^{i,j}$, where $\rvp^{i,j}$ is a learned query initialization dependent on patch location, $\overline{\rvx}$ denotes the mean body model parameters, of dimension $D'$ as in previous works~\cite{goel2023humans4d,kolotouros2019spin},
and $\oplus$ denotes concatenation along the channel axis.
Given $N$ detections, the queries $\{\rvq_n\}_{n}$ are stacked into $\rmQ^0 \in \mathbb{R}^{(D+D')\times N}$ for efficient processing in parallel.
The full feature tensor $\rmE$ is used as cross-attention keys and values.
The queries are then updated with a stack of $L$ blocks $\rmB^l$ (L=2 in practice), alternating between cross-attention layers ($\rmC\rmA$) over queries and image features, self-attention layers ($\rmS\rmA$) over queries, and an MLP:
\begin{equation}
\rmQ^l = \rmB^l \left[\rmQ^{l-1}, \rmE \right] = {\rmM\rmL\rmP}^l\left(\rmS \rmA^l \left( \rmC \rmA^l \left[\rmQ^{l-1}, \rmE \right]\right)\right).
 \label{eq:hph}
\end{equation}
The final outputs of the cross-attention-based module
are given by $\rmQ^{L} \in \mathbb{R}^{(D+D')\times N}$ and viewed as a set of $N$ output features, used to regress $N$ human-centered whole-body parameters $\big\{\rvx_n\big\}_{n}$ with a shared MLP. 

\myparagraph{Depth parametrization} 
Following the monocular depth literature~\cite{mertan2022single,weinzaepfel2023croco}, we predict the depth $d$ in log-space, also called \textit{nearness}~\cite{rajasegaran2022tracking} denoted $\eta$.
We assume a \textit{standard} focal length $\hat{f}$ and regress a \textit{normalized}  $\hat{\eta}$ 
from $\rmQ^{L}$ with an MLP:
\begin{equation} \label{eq:depth}
\rvn =\frac{\hat{f}}{f} \hat{\eta}, \hspace{0.3cm} \rvd = \text{exp}(-\eta).
\end{equation}
This follows~\cite{facil2019cam} which shows that this parametrization improves robustness to focal length changes.
The depth $d$ is used to back-project the 2D camera coordinates $\rvc$ using the camera inverse projection operator $\pi_\rmK^{-1}$ following Equation~\ref{eq:proj} to obtain the 3D location $\rvt$ of the primary keypoint.

Note that we directly supervise the \emph{absolute} depth while most  previous works~\cite{bev} supervise the \emph{relative} depth.
This is made possible by the utilization of large-scale synthetic data, where absolute depth is known, as opposed to real-world data where only relative depth can be annotated.
Our experimental results show the effectiveness of this simple strategy.

\myparagraph{Optional camera embedding}
If available, camera intrinsics $\rmK$ can be used as additional input to our model $\mathcal{H}$ which becomes $\mathcal{H}(\rmI,\rmK)$.
In more details, camera information may be integrated into the Human Perception Head at training and/or inference time.
This is a desirable feature, 
but making it optional allows for i) processing images when it is not available, and ii) fairly comparing to the state-of-the-art methods that do no use this information.

We embed camera information by computing the ray direction~\cite{mildenhall2020nerf} $\rvr_{i,j} = \rmK^{-1} [u_i,v_j,1]^T$ from each patch center  $(u_i,v_j)$.
The  first two coordinates of the $\rvr_{i,j}$ vector are kept, and embedded into a high-dimensional space using Fourier encoding~\cite{mildenhall2020nerf} to obtain a patch-level embedding $\rmE_{\rmK} \in \mathbb{R}^{H/P \times W/P \times 2(F+1)}$, where $F$ denotes the number of frequency bands.
We concatenate features extracted using the vision backbone with camera embeddings to get $\rmE \coloneqq \rmE \oplus \rmE_{\rmK}$.

\subsection{Training \Ours}
\label{sub:losses}

\Ours is fully-differentiable and trained end-to-end by back-propagation. 
We now discuss training losses.
The symbol $\sim$ denotes ground-truth targets.

\vspace{0.1cm}
\myparagraph{Detection loss}
We project the ground-truth primary keypoint of each human present in the image using the camera projection operator $\pi_\rmK$,
and construct a score map $\tilde{\rmS}$ of dimension $(W/P) \times (H/P)$ with $\tilde{s}^{i,j}$ equal to $1$ if a primary keypoint is projected to the corresponding 
patch and $0$ otherwise.
Predictions are trained by minimizing a binary cross-entropy loss: 
\begin{equation}
\mathcal{L}_{\texttt{det}} = -  \sum_{i,j} \tilde{s}^{i,j} \log(s^{i,j})  + (1 - \tilde{s}^{i,j})\log(1 - s^{i,j}).
\end{equation}

\myparagraph{Regression losses}
All other quantities predicted by the model are trained with $L_1$ regression losses. We concatenate
 the offset from the patch centers $\tilde{\rvc}$, 
 the body model parameters (pose, shape, expression) $\tilde{\rvx}$,  following~\cite{kolotouros2019spin,goel2023humans4d}, and the depth $\tilde{\rvd}$ and minimize $\mathcal{L}_{\texttt{params}} = \sum_{n} \Big| \big[\rvc, \rvx, \rvd\big] - \big[\tilde{\rvc}, \tilde{\rvx}, \tilde{\rvd}\big]\Big|.$
 We also found it beneficial to minimize an $L_1$ loss for human-centered output meshes $\mathcal{L}_{\texttt{mesh}} = \sum_{n} \big| \rmM_n - \tilde{\rmM}_n \big|$, 
as well as for the reprojection of the mesh onto the image plane $\mathcal{L}_{\texttt{reproj}} = \sum_{n} \big| \pi_\rmK(\rmM_n+\rvt_n) - \pi_\rmK(\tilde{\rmM}_n+\tilde{\rvt}_n
)|$.
The final training loss is thus:
\begin{equation}
\mathcal{L} = \mathcal{L}_{\texttt{det}} + \mathcal{L}_{\texttt{params}} + \lambda (\mathcal{L}_{\texttt{mesh}} + \mathcal{L}_{\texttt{reproj}}).
\end{equation}

\myparagraph{Synthetic whole-body \dataset dataset}
We introduce \dataset \footnote{\url{https://download.europe.naverlabs.com/ComputerVision/MultiHMR/CUFFS}}, the \datasetname dataset, designed to contain synthetic renderings of people with close-up views of full-bodies with clearly visible hands in diverse poses, see Figure~\ref{fig:cuffs}.
Using Blender~\cite{blender}, we render synthetic human models close to the camera, in poses sampled from the BEDLAM~\cite{bedlam}, AGORA~\cite{agora}, and UBody~\cite{osx} datasets, using additional hand poses from InterHand2.6M~\cite{interhand} for increased diversity.
Please refer to the supplementary material for more details.
We render a total of 60,000 images.
Simply adding this data during training improves the quality of hand pose predictions, without degrading other metrics.

\myparagraph{Implementation details} 
By default, we use squared input images of resolution $448{\times}448$, with the longest side resized to $448$ and the smallest zero-padded to maintain aspect ratio.
We use random horizontal flipping as data augmentation.
We initialize the weights of the backbone with DINOv2~\cite{dinov2} and experiment with Small, Base and Large ViT models as encoder. Please refer to the supplementary material 
for the full list of hyper-parameters and more implementation details.

%% file: sections/04_expe.tex
\section{Experiments}
\label{sec:expe}

We first ablate training data and model architecture (Section~\ref{sec:model_data}), and then compare to the state of the art on body-only and whole-body HMR (Section~\ref{sec:sota}).

\vspace{0.1cm}
\myparagraph{Evaluation metrics} 
We evaluate the accuracy of the entire 3D mesh predictions with the per-vertex error (PVE), following~\cite{romp,bev,osx}, and also report it for specific body parts (hands and face).
When the entire ground-truth mesh is not available, we report the Mean Per Joint Position Error (MPJPE) and the Percentage of Correct Keypoints (PCK) using a threshold of 15cm.
We also report these metrics after Procrustes-Alignment (PA),
and F1-Scores to evaluate detection.
To evaluate the placement in the scene, we report the Mean Root Position Error (MRPE)~\cite{bev} and the Percentage of Correct Ordinal Depth (PCOD)~\cite{zhen2020smap} metrics.
For computational costs, we report inference time on a NVIDIA V100 GPU and the number of Multiply-Add Cumulation (MACs) using the \emph{fvcore} library\footnote{\url{https://github.com/facebookresearch/fvcore}}.
More details about the metrics are given in the supplementary material.

\myparagraph{Evaluation benchmarks}
For body-only benchmarks, we predict SMPL meshes from SMPL-X meshes using the regressor from~\cite{bedlam},
and follow prior work~\cite{romp,bev,psvt,hand4whole,osx} in evaluating on 3DPW~\cite{3dpw}, MuPoTs~\cite{mupots}, CMU~\cite{cmu} and AGORA~\cite{agora}.
For whole-body evaluation, we compare performance with prior work~\cite{osx,pixie,hand4whole} on EHF~\cite{ehf}, AGORA~\cite{agora} and UBody~\cite{osx}.
We refer to the supplementary material for more details on datasets.

\input{tab/12_model_and_data}
\input{tab/10_hph}

\subsection{Ablations on model design and training data}
\label{sec:model_data}

\myparagraph{Default configuration}
For the ablations, we use a ViT-B backbone with a HPH head composed of $2$ blocks. 
We train only using synthetic the BEDLAM and AGORA datasets (but not \dataset), without using the intrinsics as input.
In each table the row of the default ablation configuration has a grey background.

\myparagraph{Model architecture}
We investigate several architectures in Table~\ref{tab:arch}.
As most state-of-the-art \onestage methods (ROMP~\cite{romp}, BEV~\cite{bev}, PSVT~\cite{psvt}) use a HRNet~\cite{hrnet} convolutional backbone, we evaluate both HRNet and ViT-S 
(as they have approximately equivalent parameter counts, 28.6M for HRNet and 21M for ViT-S)
with either a vanilla iterative regression head~\cite{kanazawa2018hmr} (`Reg.') or our proposed HPH.
In both cases, the ViT-S backbone is beneficial and significant gains also come from our proposed HPH head, which validates our architecture.
Scaling up the backbone (last row) further improves performance. 

\myparagraph{Training data}
In Table~\ref{tab:data}, we experiment with different types of training data. One source can be real-world datasets (`Real': MS-CoCo~\cite{coco}, MPII~\cite{mpii} and Human3.6M~\cite{h36m}), for which pseudo-ground-truth fits~\cite{Moon_2022_CVPRW_NeuralAnnot,Moon_2023_CVPRW_3Dpseudpgts} are obtained by minimizing the reprojection error of annotated 2D keypoints, but this remains inherently noisy.
An alternative is to train on synthetic datasets such as AGORA~\cite{agora} (`A') or BEDLAM~\cite{bedlam} (`B') that have the advantage to be highly scalable and to have perfect ground-truth. Recent work~\cite{bedlam} has shown that state-of-the-art results can be achieved using synthetic training data only,
despite an inherent sim-to-real gap.
Our results confirm this finding as we obtain better results when training on large-scale synthetic data.
When we add our synthetic \dataset dataset (`C') we observe a significant boost in performance especially for metrics related to the hands (column EHF-H in the fourth row).
However, when combining both real-world and synthetic datasets (last row), performance drops compared to training solely on synthetic data (penultimate row).

\myparagraph{HPH architecture} In Table~\ref{tab:hph}, we further compare different heads to regress the SMPL-X parameters.
The baseline (`Reg.') uses a vanilla iterative regressor~\cite{kolotouros2019spin} applied to each detected feature token independently.
`HPH' converges faster (Table~\ref{tab:hph_converg}) and performs better (Table~\ref{tab:hph_arch}).
`HPH~w/o~$\rmS\rmA$' denotes a variant where queries are treated independently by removing $\rmS\rmA$ blocks from the HPH, see Equation~\ref{eq:hph}: treating queries together is beneficial (Table~\ref{tab:hph_arch}).
In Table \ref{tab:hph_hyperparams} we experiment with different configurations of the HPH (number of layers `L' and number of attention heads `H').
Increasing the number of layers slightly improves performance but we favor the use of 2 layers for better efficiency.

\input{tab/08_effcurve}

\myparagraph{Input resolution and backbone size} We 
evaluate the impact of the input image resolution 
for different backbone sizes (ViT-S, ViT-B, ViT-L) in Figure~\ref{fig:resolution}. Increasing the input resolution consistently brings performance gains across backbone sizes, at the cost of increased inference time (right). For body-only metrics, a ViT-L backbone at $448{\times}448$ inputs arguably offers a good performance \emph{vs.} speed trade-off. Using higher resolutions may be more worthwhile for whole-body metrics; in particular, with a ViT-S or ViT-B backbone, high resolutions are critical to achieve competitive performance.
This is to be expected as small details such as facial expressions and hand poses are easier to capture at high resolution -- it motivated previous works~\cite{choutas2020expose, pixie, hand4whole} to extract specific high resolution crops for these parts.
The largest backbone (ViT-L) at a $896{\times}896$ resolution takes approximately $120$ms per image -- without compressing or quantizing the network -- which is fast compared to multi-stage methods (see Section~\ref{sec:sota}).

\myparagraph{Optional camera intrinsics} Integrating camera information is expected to improve accuracy when recovering and placing human 3D meshes in the scene.
In Table~\ref{tab:camera}, we report results with different kinds of camera embeddings: computing \textit{simple} embedding (where the flattened intrinsics matrix is fed to a linear layer) degrades performances compared to not adding camera embedding (\ie, \textit{none}) while adding \textit{rays} brings a gain.
When combined with focal length normalization $\hat{f}$, we observe a clear gain on all metrics.
In Table~\ref{tab:intrisics} we report:
performance with a fixed field of view (FOV) of $60^\circ$, like ROMP/BEV, for a model trained with intrinsics (row 1), and for a model trained without (row 2). Conditioning the model on camera intrinsics improves depth prediction accuracy (row 3), while reconstruction metrics which are centered on people are far less sensitive to this change.
This validates the benefit of using intrinsics when available.

\input{tab/01_ablations}

\begin{figure*}[t]
 \includegraphics[width=\textwidth]{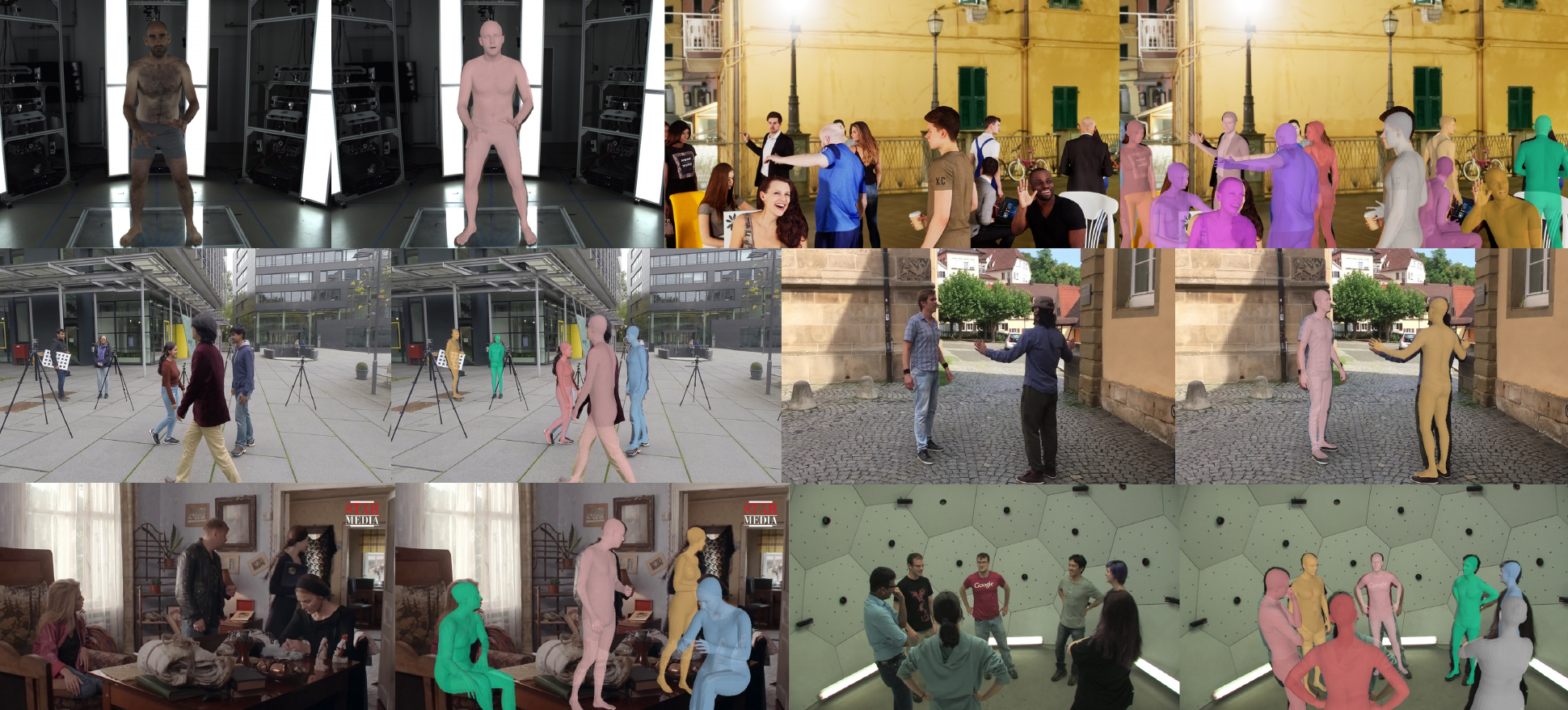}\\[-0.28cm]
 \caption{\textbf{Randomly sampled qualitative examples:} input image and our results overlaid on it.
 Images from EHF and AGORA (top), MuPoTs and 3DPW (middle), UBody and CMU (bottom). See supplementary material for more visualizations. 
 }
 \label{fig:qualitative}
\end{figure*}

\myparagraph{Other design choices} We present other ablations, \eg on training losses and choice of primary keypoints, in the supplementary material.

\myparagraph{Qualitative results} Figure~\ref{fig:qualitative} shows visualizations of some predictions.

\input{tab/00_sota}

\subsection{Comparisons with the state of the art}
\label{sec:sota}

No existing method is both multi-person and whole-body (Table~\ref{tab:feature}).
We thus compare either to multi-person approaches on body-only mesh recovery or to whole-body methods. 
In the latter case, our approach is single-shot, while others assume human detections, extract crops around each person, and process each one independently.
We report results with a $896{\times}896$ input resolution and without using  camera intrinsics, 
with either a model finetuned for each benchmark as other methods do or a single universal model indicated by $\dagger$ (please refer to the supplementary material for additional information regarding finetuning).

\input{tab/13_pretraining_depth_macs}

\myparagraph{Body Mesh Recovery}
As most of these methods (ROMP~\cite{romp}, BEV~\cite{bev} and PSVT~\cite{psvt}) use a $512{\times}512$ resolution, we also report results obtained at $448{\times}448$, which offers an excellent speed-performance trade-off.
All these multi-person approaches are limited to body-only meshes.
\Ours outperforms existing work, with substantial gains across various metrics, even when using lower resolution input, smaller backbone or a universal model.
At the same time, it also predicts hands poses and facial expressions (as evaluated next), which is not the case for other multi-person approaches.

\myparagraph{Whole-Body Mesh Recovery}
We evaluate our whole-body regression performance by comparing it against whole-body 3D pose methods~\cite{osx,pixie,hand4whole}. 
All existing approaches are limited to the single-person scenario: they do not consider the detection stage and the 3D positions in the scene, instead assuming predefined 2D bounding boxes around the person of interest.
We report results in Table~\ref{tab:sota_smplx}.
\Ours is competitive with, or outperforms, previous whole-body methods, even when considering the universal model.
In particular it obtains competitive performance on hands and faces (on par with or better than SMPLer-X~\cite{cai2023smplerx}, that is not single-shot).
Overall, empirical results show that \Ours predicts accurate hand and facial poses while also being multi-person.

\myparagraph{Human depth estimation} In Table~\ref{tab:depth}, we compare the performance of our model in distance estimation, which uses simple depth regression, to the state of the art~\cite{romp,bev,mehta2020xnect}.
Prior works assume a fixed camera setting. For example, BEV~\cite{bev} is competitive on AGORA-val but does not generalize as well to datasets with different camera parameters.
The camera-aware variant of \Ours provides accurate distance predictions across datasets and camera parameters, and the proposed approach still significantly outperforms the state of the art when camera intrinsics are not provided.

\myparagraph{Inference cost}
The number $N$ of humans in an image defines the number of queries in the HPH head.
With $N{=}512$, HPH takes 2.5ms \emph{vs.} 2.3ms for $N{=}5$ on a NVIDIA V100 GPU.
Other parts of the model are independent of $N$, thus our method scales well, as do other single-shot approaches (\eg ROMP, BEV), see Table~\ref{tab:macs}.
This is in contrast to multi-stage methods (\eg Hand4Whole, OSX) which detect people,\eg with YOLOv5~\cite{yolov5}, and independently process their crops.

%% file: tab/12_model_and_data.tex
\begin{table*}[t]
\caption{
\textbf{Architecture and training data} are ablated on MuPoTs (PCK3D-All), 3DPW (MPJPE), EHF (PVE-All), EHF-H (PVE-Hands) and CMU (MPJPE).
Default settings in \colorbox{baselinecolor}{grey}.
\textbf{(a)} We compare a ViT backbone to HRNet as well as our HPH with respect to a standard iterative regressor~\cite{kanazawa2018hmr} (`Reg.').
\textbf{(b)} Training data type; `Real'=MS-CoCo+MPII+Human3.6M,`A'=AGORA, `B'=BEDLAM, and `C'=\dataset. When trained on `C' only, we evaluate on single-person test sets only.
\\[-0.6cm]
}
\subfloat[
\textbf{Architecture} 
\label{tab:arch}
]{
\begin{minipage}{0.48\linewidth}
\begin{tabular}{y{32}y{20}@{\hspace{0.05cm}}|x{32}x{28}x{20}x{20}}
{\scriptsize Backbone} & {\scriptsize Head}  & {\scriptsize MuPoTs$\uparrow$} & {\scriptsize 3DPW$\downarrow$} & \scriptsize {EHF$\downarrow$} & {\scriptsize CMU$\downarrow$} \\
\shline
HRNet  & Reg. & 65.8 & 83.2 & 143.1 & 130.1 \\
ViT-S  & Reg. & 70.1 & 80.2 & 90.6 & 118.1 \\
\hline
HRNet  & HPH & 69.8 & 80.2 & 115.2 & 116.6 \\
ViT-S  & HPH & 70.9 & 80.1 & 80.1 & 109.1 \\
\baseline{ViT-B} & \baseline{HPH} & \baseline{\textbf{76.3}} & \baseline{\bf{73.5}} & \baseline{\textbf{55.3}} & \baseline{\textbf{97.2}} \\
\end{tabular}
\end{minipage}
}
\hfill
\subfloat[
\textbf{Data} 
\label{tab:data}
]{
\begin{minipage}{0.48\linewidth}
\begin{tabular}{y{34}|x{34}x{26}x{16}x{34}@{\hspace{-0.05cm}}x{20}}
 {\scriptsize} Data & {\scriptsize MuPoTS$\uparrow$} & {\scriptsize 3DPW$\downarrow$} & {\scriptsize EHF$\downarrow$} & {\scriptsize EHF-H$\downarrow$} & {\scriptsize CMU$\downarrow$} \\
\shline
Real & 68.5  & 83.8 & 70.2 & 51.2 & 101.6 \\
\baseline{A+B}  & \baseline{\bf{76.3}} & \baseline{73.5} & \baseline{55.3} & \baseline{47.4} & \baseline{97.2} \\
C & - & - & 53.5 & 44.5 & - \\
A+B+C & 76.0 & \bf{72.9} & \bf{49.8} & \bf{40.5} & \bf{96.5}\\
~~+Real & 69.8 & 77.6 & 61.1 & 48.4 & 98.5 \\
\arrayrulecolor{white}\hline
\end{tabular}
\end{minipage}
}
\end{table*}

%% file: tab/10_hph.tex
\begin{table*}[!b]
\label{tab:hph_conf}
    \caption{
    \textbf{Ablation on the Human Perception Head (HPH).} 
    `Reg.': parallel iterative regressors; HPH w/o $\rmS\rmA$: queries processed independently in HPH, \ie, without self-attention, 
    $L$: number of layers and $H$: number of heads.
    \textbf{(a)} Training convergence speed.
    \textbf{(b)} Impact of head choice.
    \textbf{(c)} Impact of HPH hyperparameters.
    \\[-0.6cm]
    }
\label{tab:hph}
\subfloat[
\textbf{Convergence} 
\label{tab:hph_converg}
]{
\hspace*{-0.4cm}
\begin{minipage}{0.32\linewidth}
\resizebox{!}{0.5\linewidth}{
    \begin{tikzpicture}
    \begin{axis}[
        width=1.4\linewidth,
        height=3.5cm,
        ylabel= {\scriptsize MuPoTs \\ PCK3D ($\uparrow$)},
        ylabel shift = -0.2cm,
        xlabel = Iterations,
        xlabel shift = -0.1cm,
        xmin=0, xmax=300000,
        ylabel style={align=center, font=\scriptsize},
        xlabel style={align=center, font=\scriptsize},
        tick label style={font=\scriptsize},  
        xticklabels = {0, 100k, 200k, 300k},
        every axis plot/.append style={ultra thick},
        legend pos=south east,
        x tick label style={yshift={0.05cm}},
        scaled x ticks = false,
        legend style = {column sep=1pt, font=\scriptsize, only marks, opacity=1},
        ]
        \addplot[style=NavyBlue] coordinates {
                (4000 , 59.8)
                (16000, 66.0)
                (28000, 67.8)
                (40000, 69.9)
                (52000, 71.8)
                (64000, 72.8)
                (76000, 73.5)
                (88000, 73.5)
                (104000, 75.0)
                (160000, 75.4)
                (220000, 76.0)
                (304000, 76.3)
                };
        \addlegendentry{HPH}
        \addplot[style=black] coordinates {
                (4000 , 54.7)
                (16000, 61.7)
                (28000, 63.4)
                (40000, 64.0)
                (52000, 64.3)
                (64000, 66.1)
                (76000, 66.9)
                (88000, 67.3)
                (100000, 67.8)
                (160000, 70.3)
                (220000, 72.2)
                (304000, 73.5)
                };
        \addlegendentry{Reg.}
    \end{axis}
    \end{tikzpicture}
}
\end{minipage}
}
\hfill
\subfloat[
\textbf{Head architecture} 
\label{tab:hph_arch}
]{
\hspace*{-0.5cm}
\begin{minipage}{0.35\linewidth}
\flushleft
\resizebox{1.1\linewidth}{!}{
\renewcommand{\arraystretch}{1.2}
\begin{tabular}{y{50}|x{40}x{28}x{30}} % 
  Head & {MuPoTS$\uparrow$} & {3DPW$\downarrow$} & {EHF$\downarrow$} \\
    \shline
    Reg.~\cite{kanazawa2018hmr} & 73.5 & 78.9 & 65.0 \\
    HPH \textsubscript{w/o $\rmS\rmA$} & 74.5 & 76.4 & 63.2 \\
    HPH & \baseline{\bf{76.3}} &  \baseline{\bf{73.5}} & \baseline{\bf{55.3}} \\
    \multicolumn{4}{c}{~}\\
\end{tabular}
}
\end{minipage}
}
\hfill
\subfloat[
\vspace*{-0.25cm}
\textbf{HPH Hyperparameters} 
\label{tab:hph_hyperparams}
]{
\begin{minipage}{0.29\linewidth}
\flushright
\resizebox{\linewidth}{!}{
\begin{tabular}{x{6}x{10}|x{38}x{28}x{26}}
    $L$ & $H$ & {MuPoTS$\uparrow$} & {3DPW$\downarrow$} & {EHF$\downarrow$} \\
    \shline
    \baseline{2} &  \baseline{8} &  \baseline{76.3} & \baseline{73.5}  & \baseline{55.3} \\
    2 &  4 & 76.5 & 74.0 & 54.8 \\
    4 &  8 &  78.5 & 72.4  & 51.3 \\
    8 &  8 &  \bf{78.9} & \bf{72.0}  & \bf{51.0} \\
\end{tabular}
}
\end{minipage}
}
\end{table*}

%% file: tab/08_effcurve.tex
\begin{figure*}[t!]
\begin{center}
\resizebox{0.45\linewidth}{!}
{
    \begin{tikzpicture}[baseline=5cm]
    \begin{axis}[
        hide axis,
        xmin=10, xmax=50,
        ymin=0,ymax=0.4,
        legend columns=4,
        legend style={draw=white!15!black, /tikz/every even column/.append style={column sep=0.3cm}},
        ]
        \addlegendimage{bbdinovtwos} 
        \addlegendentry{ViT-S}
        \addlegendimage{bbdinovtwob} 
        \addlegendentry{ViT-B}
        \addlegendimage{bbdinovtwol} 
        \addlegendentry{ViT-L}
        \addlegendimage{sotaline} 
        \addlegendentry{SotA}
    \end{axis}
    \end{tikzpicture}
    }\\ 
\vspace*{-2.5cm}
\begin{tikzpicture}
\begin{axis}[
    width=0.29\linewidth,
    height=3.5cm,
    ylabel= MuPoTs - PCK3D ($\uparrow$),
    xlabel = Resolution,
    xlabel shift=-0.15cm,
    ylabel shift=-0.2cm,
    ylabel style={align=center, font=\tiny},
    xlabel style={align=center, font=\tiny},
    tick label style={font=\tiny},
    xmin=200, xmax=920,
    every axis plot/.append style={ultra thick},
    legend pos=south east,
    legend style = {column sep=2pt, font=\tiny, only marks},
]
    \addplot[style=bbdinovtwos] coordinates {
            (224, 60.3) 
            (448, 70.9)
            (504, 71.3)
            (672, 72)
            (896, 73.5)
            };
    \addplot[style=bbdinovtwob] coordinates {
            (224, 63.9)
            (448, 74.3)
            (504, 78.1)
            (672, 79.0)
            (896, 80.1)
            };
    \addplot[style=bbdinovtwol] coordinates {
            (224, 69.4)
            (448, 80.6)
            (504, 82.0)
            (672, 83.3)
            (896, 83.6)
            };
    \draw[color=ForestGreen, line width=0.8pt,dashed]  (170, 70.2) -- (950, 70.2);
    \addplot[style=sotastyle] coordinates {(512, 70.2)};
\end{axis}
\end{tikzpicture}
\hfill
\begin{tikzpicture}
\begin{axis}[
    width=0.29\linewidth,
    height=3.5cm,
    ylabel= CMU - MPJPE ($\downarrow$),
    ylabel shift=-0.2cm,
    xlabel = Resolution,
    xlabel shift=-0.15cm,
    ylabel style={align=center, font=\tiny},
    xlabel style={align=center, font=\tiny},
    tick label style={font=\tiny},
    xmin=200, xmax=920,
    every axis plot/.append style={ultra thick},
    legend pos=north east,
    legend style = {column sep=2pt, font=\tiny, only marks},
]
    \addplot[style=bbdinovtwos] coordinates {
            (224, 125) 
            (448, 108)
            (504, 107)
            (672, 103)
            (896, 101)
            };
    \addplot[style=bbdinovtwob] coordinates {
            (224, 122)
            (448, 98)
            (504, 93)
            (672, 88)
            (896, 90)
            };
    \addplot[style=bbdinovtwol] coordinates {
            (224, 105)
            (448, 84)
            (504, 83)
             (672, 81.3)
             (896, 81)
            };
    \draw[color=ForestGreen, line width=0.8pt,dashed]  (170, 105.7) -- (950, 105.7);
    \addplot[style=sotastyle] coordinates {(512, 105.7)};
\end{axis}
\end{tikzpicture}
\hfill
\begin{tikzpicture}
\begin{axis}[
    width=0.29\linewidth,
    height=3.5cm,
    ylabel shift=-0.2cm,
    ylabel= EHF - MVE($\downarrow$),
    xlabel = Resolution,
    xlabel shift=-0.2cm,
    ylabel style={align=center, font=\tiny},
    xlabel style={align=center, font=\tiny},
    tick label style={font=\tiny},
    xmin=200, xmax=920,
    every axis plot/.append style={ultra thick},
    legend pos=north east,
    legend style = {column sep=2pt, font=\tiny, only marks},
]
    \addplot[style=bbdinovtwos] coordinates {
              (224, 113)
            (448, 83)
            (504, 80)
            (672, 73)
            (896, 73)
            };
    \addplot[style=bbdinovtwob] coordinates {
           (224, 90) 
            (448, 77)
            (504, 74)
            (672, 63)
            (896, 62.5)
            };
    \addplot[style=bbdinovtwol] coordinates {
            (224, 81)
            (448, 65.1)
            (504, 62)
            (672, 60)
            (896, 55.2)
            };
    \draw[color=ForestGreen,  line width=0.8pt,dashed]  (170, 70.8) -- (950, 70.8);
    \addplot[style=sotastyle] coordinates {(800, 70.8)};
\end{axis}
\end{tikzpicture}
\hfill
\begin{tikzpicture}
\begin{axis}[
    width=0.35\linewidth,
    height=3.5cm,
    ymode=log,
    ylabel= Inference time (ms),
    ylabel shift=-0.2cm,
    xlabel = Resolution,
    xlabel shift=-0.2cm,
    ylabel style={align=center, font=\tiny},
    xlabel style={align=center, font=\tiny},
    tick label style={font=\tiny},
    xmin=200, xmax=920,
    ytick = {20, 30, 50, 100},
    yticklabels = {20, 33, 50, 100},
    every axis plot/.append style={ultra thick},
    legend pos=north west,
    legend style = {column sep=2pt, font=\tiny, only marks},
]
    \addplot[style=bbdinovtwos] coordinates {
            (224, 20) 
            (448, 26)
            (672, 29)
            (896, 38)
            };
    \addplot[style=bbdinovtwob] coordinates {
            (224, 24)
            (448, 36)
            (672, 43)
            (896, 59)
            };
    \addplot[style=bbdinovtwol] coordinates {
            (224, 34)
            (448, 43)
            (672, 74)
            (896, 127)
            };
\end{axis}
\end{tikzpicture}
\\[-0.7cm]
\caption{\textbf{Backbone-resolution-speed trade-off.} We report the performance on MuPoTs, CMU and EHF using different backbone sizes and image resolutions. We also report the inference time (right). 
}
\label{fig:resolution}
\end{center}
\end{figure*}

%% file: tab/01_ablations.tex
\begin{table*}[t]
\caption{
\textbf{Ablative study}.
{
Experiments on \textbf{(a)} the importance of the camera embedding type and \textbf{(b)} the sensitivity to the camera intrinsics in terms of human-centric reconstruction error and distance estimation error.
$\hat{f}$: focal length normalization.
\\[-0.6cm]
}
\label{tab:ablations}
}
\centering
\begin{minipage}{0.31\linewidth}
\centering
\subcaption{\textbf{Camera embeddings} \label{tab:camera}}
\vspace{-0.08cm}
\scalebox{0.8}{
\begin{tabular}{y{28}x{38}x{28}x{28}}
 & {\scriptsize MuPoTS$\uparrow$} & {\scriptsize 3DPW$\downarrow$} & {\scriptsize EHF$\downarrow$}\\
\shline
\baseline{none} & \baseline{76.3} & \baseline{73.5} & \baseline{55.3}\\
simple & 74.8 & 75.3& 56.8 \\
rays &  77.0 & 72.6 & 54.4 \\
rays+$\hat{f}$ & \bf{78.8} & \bf{71.3} & \bf{53.1} \\
\end{tabular}
}
\end{minipage}
\hfill
\begin{minipage}{0.65\linewidth}
\centering
\subcaption{\textbf{Impact of optional intrinsics} \label{tab:intrisics}}
\vspace{-0.08cm}
\scalebox{0.8}{
\setlength{\tabcolsep}{4pt}
\begin{tabular}{cc|ccc|ccc}
 \multicolumn{2}{c|}{FOV}  & \multicolumn{3}{c|}{Reconstruction$\downarrow$} & \multicolumn{3}{c}{Distance (MRPE$\downarrow$)} \\
\shline
Train & Test & {\small MuPoTs} & {\small 3DPW} & {\small CMU} & {\small MuPoTs} & {\small 3DPW} & {\small CMU} \\
\hline
\baseline{$60^\circ$} & \baseline{$60^\circ$} & \baseline{76.3}  & \baseline{73.5} & \baseline{97.2}& \baseline{1345 }  & \baseline{732} & \baseline{570}\\
gt & $60^\circ$ & 76.8  & 76.8 & 99.5& 1512  & 731 & 595\\
gt & gt & \bf{76.5}  & \bf{73.2} & \bf{96.9}& \bf{693} 
& \bf{445} & \bf{287}\\
\end{tabular}
}
\end{minipage}
\end{table*}

%% file: tab/00_sota.tex
\begin{table*}[thb]
\vspace{.2em}
\centering
\subfloat[
\textbf{Body-only benchmarks}
\label{tab:sota_smpl}
]{
\hspace{-1em}
\centering
\resizebox{\linewidth}{!}{
\begin{tabular}{l@{\hspace{-0.1em}}ccc|ccc|cc|cc|ccc}
\multirow{2}{*}{\bf{Method}} & \multirow{2}{*}{\bf{Res.}} & \bf{Single} & \multirow{2}{*}{\bf{Backbone}} & \multicolumn{3}{c|}{\bf{3DPW}} & \multicolumn{2}{c|}{\bf{MuPoTs}} & \multicolumn{2}{c|}{\textbf{CMU}} & \multicolumn{3}{c}{\bf{AGORA}} \\
& & \bf{Shot} & &  PA-MPJPE↓ & MPJPE↓ & PVE↓ & PCK-All↑ & PCK-Matched↑ & F1↑ & MPJPE↓ &  F1↑ & MPJPE↓ & PVE↓ \\
\shline
\rowcolor{gray!10} \textit{Body-only} &&&&&&&&&&&&& \\
CRMH~\cite{crmh}       & 832 & \checkmark & RN50 & -    & -    & -    & 69.1 & 72.2 & 0.92 & 143.2 & -         & -     & - \\
3DCrowdNet~\cite{3dcrowdnet} & Full &            & RN50 & 51.5 & 81.7 & 98.3 & 72.7 & 73.3 & 0.95 & 127.3 & -         & -     & \\
ROMP~\cite{romp}       & 512 & \checkmark & HR32 & 47.3 & 76.6 & 93.4 & 69.9 & 72.2 & 0.93 & 128.2 & 0.91      & 108.1 & 103.4 \\
BEV~\cite{bev}         & 512 & \checkmark & HR32 & 46.9 & 78.5 & 92.3 & 70.2 & 75.2 & \bf{0.97} & 109.5 & 0.93 & 105.3 & 100.7 \\
PSVT~\cite{psvt}        & 512 & \checkmark & HR32 & 45.7 & 75.5 & 84.9 & -    & -    & \bf{0.97} & 105.7 & 0.93 & 97.7  & 94.1 \\
\shline
\rowcolor{gray!10} \textit{Whole-Body} &&&&&&&&&&&&& \\
Hand4Whole~\cite{hand4whole}        & Full &            & RN50     & 54.4       & 86.6       & -          & -          & -          & - & -         & 0.93 & \underline{89.8}      & \underline{84.8}  \\
OSX~\cite{osx}                      & Full &            & ViT-L/16 & 60.6       & 86.2       & -          & -          & -          & - & -         & -         & -         & - \\
SMPLer-X~\cite{cai2023smplerx}      & Full &            & ViT-L/16 & 51.5       & 76.8       & -          & -          & -          & - & -         & -         & -         & -  \\
SMPLer-X~\cite{cai2023smplerx}      & Full &            & ViT-H/16 & 48.0       & 71.7       & -          & -          & -          & - & -         & -         & -         & - \\
\rowcolor{NavyBlue!10} \bf{\Ours}           & 896 & \checkmark & ViT-S/14 & 53.2 & 76.3 & 91.1 & 77.0       & 81.5       & \bf{0.97} & 102.9     & -         & -         & -  \\
\rowcolor{NavyBlue!10} \bf{\Ours}           & 896 & \checkmark & ViT-B/14 & 46.7 & 70.9 & 86.9 & 79.4       & 84.6       & \bf{0.97} & 94.6      & -         & -         & - \\
\rowcolor{NavyBlue!10} \bf{\Ours}           & 896 & \checkmark & ViT-L/14 & \bf{41.7}  & \bf{61.4}  & \bf{75.9}  & \bf{85.0} & \bf{89.3} & \bf{0.97} & \bf{77.3} & \bf{0.95} & \bf{65.3} & \bf{61.1}\\
\rowcolor{NavyBlue!10} \bf{\Ours}           & 448 & \checkmark & ViT-L/14 & \underline{43.8} & \underline{64.6} & \underline{79.7} & 77.8 & 84.1 & \underline{0.96} & \underline{84.0} & - & - & -  \\
\rowcolor{NavyBlue!10} \bf{\Ours}$^\dagger$ & 896 & \checkmark & ViT-L/14 & 46.9       & 69.5       & 88.8       & \underline{80.6}      & \underline{86.4}       & \bf{0.97} & 97.5      & -         & -         & - \\
\end{tabular}
}
}
\\
\centering
\vspace{.4em}
\subfloat[
\textbf{Whole-body benchmarks}
\label{tab:sota_smplx}
]{
\hspace{-1em}
\centering
\resizebox{\linewidth}{!}{
\setlength{\aboverulesep}{0pt}
\setlength{\belowrulesep}{0pt}
\begin{tabular}{l@{\hspace{-0.5em}}cc|ccc|ccc|ccc|ccc|ccc}
                                                      & &  & \multicolumn{6}{c|}{\bf{EHF}} & \multicolumn{3}{c|}{\bf{AGORA}} & \multicolumn{6}{c}{\bf{UBody-intra}} \\
\multirow{2}{*}{\bf{Method}} & \bf{Single} & \multirow{2}{*}{\bf{Backbone}} & \multicolumn{3}{c|}{PVE↓}     & \multicolumn{3}{c|}{PA-PVE↓}                & \multicolumn{3}{c|}{PVE↓}               & \multicolumn{3}{c|}{PVE↓} & \multicolumn{3}{c}{PA-PVE↓} \\
& \bf{shot}   & & All & Hands & Face & All & Hands & Face & All & Hands & Face & All & Hands & Face & All & Hands & Face\\
\shline
\rowcolor{gray!10} \multicolumn{3}{l|}{\textit{Single person, per-body-part crops}}           & & & & & & & & & & & & &&&\\
ExPose~\cite{choutas2020expose}       &  & HR32/RN18    & 77.1  & 51.6 & 35.0 & 54.5 & 12.8 & 5.8 & 217.3 & 73.1 & 51.1 & -     & -    & -    & -    & -    & - \\
FrankMocap~\cite{rong2021frankmocap} &  & RN50    & 107.6 & 42.8 & -    & 57.5 & 12.6 & -   & -     & 55.2 & -    & -     & -    & -    & -    & -    & -   \\
PIXIE~\cite{pixie}                    &  & RN50    & 88.2  & 42.8 & 32.7 & 55.0 & 11.1 & 4.6 & 191.8 & 49.3 & 50.2 & 168.4 & 55.6 & 45.2 & 61.7 & 12.2 & 4.2 \\
Hand4Whole~\cite{hand4whole}          &  & RN50    & 76.8  & 39.8 & 26.1 & 50.3 & 10.8 & 5.8 & 135.5 & 47.2 & 41.6 & 104.1 & 45.7 & 27.0 & 44.8 & 8.9  & 2.8 \\
PyMAF-X~\cite{pymafx2023}             &  & HR48 & 64.9  & \underline{29.7} & 19.7 & 50.2 & \bf10.2 & 5.5 & 125.7 & \underline{45.0} & 35.0 & -     & -    & -    & -    & -    & - \\
\midrule
\rowcolor{gray!10} \multicolumn{3}{l|}{\textit{Single person, feature resampling}}             & & & & & & & & & & & &&&& \\
OSX~\cite{osx}                     &  & ViT-L/16 & 70.8 & 53.7 & 26.4 & 48.7 & 15.9 & 6.0 & 122.8 & 45.7 & 36.2 & 81.9 & 41.5 & 21.2 & 42.2 & 8.6  & \underline{2.0}\\
SMPLer-X~\cite{cai2023smplerx} &  & ViT-L/16 & 65.4 & 49.4 & \bf17.4 & 37.8 & 15.0 & \bf5.1 & \underline{99.7}  & \bf39.3 & \underline{29.9} & 57.4 & 40.2 & 21.6 & 31.9 & 10.3 & 2.8  \\
\midrule
\rowcolor{gray!10} \multicolumn{3}{l|}{\textit{Multi-person, one forward pass}}             & & & & & & & & & & & & &&&\\
\rowcolor{NavyBlue!10} \bf{\Ours} & \checkmark & ViT-S/14 & 50.0 & 43.3 & 24.4 & 36.8 & 14.4 & 5.8 & - & - & - & 56.9 & 35.7 & 18.9 & 23.8 & 9.9 & 2.5 \\
\rowcolor{NavyBlue!10} \bf{\Ours} & \checkmark & ViT-B/14 & \underline{43.3} & 39.5 & 23.3 & \underline{34.8} & 12.2 & 5.4 & - & - & - & 54.4 & 32.0 & 17.3 & 23.0 & 8.8 & 2.2 \\
\rowcolor{NavyBlue!10} \bf{\Ours} & \checkmark & ViT-L/14 & \bf42.0 & \bf28.9 & \underline{18.0} & \bf28.2 & \underline{10.8} & \underline{5.3} & \bf{95.9} & \underline{40.7} & \bf{27.7} & \bf51.2 & \bf25.0 & \bf16.2 & \bf21.0 & \bf7.2 & \bf1.8 \\
\rowcolor{NavyBlue!10} \bf{\Ours} $\dagger$ & \checkmark & ViT-L/14 & \bf42.0 & \bf28.9 & \underline{18.0} & \bf28.2 & \underline{10.8} & \underline{5.3} & - & - & - & \underline{54.0} & \underline{27.5} & \underline{17.0} & \underline{22.8} & \underline{8.0} & 2.4 \\
\end{tabular}
}
}
\caption{
\textbf{Comparison with state-of-the-art methods.}
As there is no other method that is both multi-person and whole-body, we compare separately to state-of-the-art approaches for 
\textbf{(a)} multi-person body-only mesh recovery, and 
\textbf{(b)} whole-body mesh recovery (all methods except \Ours are single-person).
For AGORA, we report performance for a single \Ours setting due to restrictions of the evaluation system.
$\dagger$ indicates a universal model which is not finetuned specifically for each benchmark. 
\vspace*{-0.3cm}
}
\label{tab:sota}
\end{table*}

%% file: tab/13_pretraining_depth_macs.tex
\begin{table*}[t]
\caption{\textbf{
Comparison to existing works for human depth estimation and inference cost.}
\textbf{(a)} Human depth estimation: we evaluate \Ours without and with camera intrinsics information.
\textbf{(b)} Comparison of inference cost for different number of humans $N$ in an image between \Ours (bottom) and the state of the art, which is limited to either multi-person but body-only methods (top), or single-person whole-body approaches thus requiring a human detector (middle).
\\[-0.6cm]
}
\label{tab:ablations_last}
\centering
\subfloat[
    \textbf{Depth estimation benchmark}
    \label{tab:depth}
]{
    \vspace{0.05cm}
    \begin{minipage}{0.42\linewidth}
    \centering
    \resizebox{\linewidth}{!}{
       \renewcommand{\arraystretch}{1.2}
      \begin{tabular}{y{57}|x{28}x{23}x{21}x{28}|x{27}x{21}}
             & \multicolumn{4}{c|}{MRPE ($\downarrow$)} & \multicolumn{2}{c}{PCOD ($\uparrow$)} \\
             {\small Method} & { \scriptsize MuPoTs} & {\scriptsize 3DPW} & {\scriptsize CMU} & {\scriptsize AGORA} & {\scriptsize MuPoTs} & {\scriptsize CMU} \\
             \hline
             XNect~\cite{mehta2020xnect} & \cellcolor{tabsecond}639 & - & - & - & - & -\\
             ROMP~\cite{romp}& 1688 & 1060 & 679 & -  & 91.2 &  \cellcolor{tabthird}97.1\\
             BEV~\cite{bev}& 1884 & \cellcolor{tabthird}1030 & \cellcolor{tabthird}673 & \cellcolor{tabthird}518  &  \cellcolor{tabthird}91.3 & 91.2\\
             \bf{\Ours} & & & & & & \\
             \bf{\small \hspace{0.1cm} w/o cam.} & \cellcolor{tabthird}1125 & \cellcolor{tabsecond}522 & \cellcolor{tabsecond}355 & \cellcolor{tabsecond}421  & \cellcolor{tabsecond}95.1 & \cellcolor{tabsecond}98.5\\  
             \bf{\small \hspace{0.1cm} w/ cam.} & \cellcolor{tabfirst}\bf{514} & \cellcolor{tabfirst}\bf{318} & \cellcolor{tabfirst}\bf{110} & \cellcolor{tabfirst}\bf{396}  & \cellcolor{tabfirst}\bf{97.9} & \cellcolor{tabfirst}\bf{99.5}\\
        \end{tabular}\hspace*{-0.3cm}  
    }
    \end{minipage}
}
\subfloat[
    \textbf{Inference time and MACs}
    \label{tab:macs}
]{
    \begin{minipage}{0.55\linewidth}{
    \centering
    \small
    \resizebox{\linewidth}{!}{
    \renewcommand{\arraystretch}{1}
    \begin{tabular}{y{68}@{\hspace{-0.2cm}}z{34}@{\hspace{0.08cm}}|c|ccc|ccc}
    \multirow{2}{*}{Method} & \multirow{2}{*}{SMPL-X} & \multicolumn{1}{c|}{Params} & \multicolumn{3}{c|}{Time (ms)} & \multicolumn{3}{c}{MACs (G)} \\
     & & \multicolumn{1}{c|}{(M)} & $N{=}1$ & $N{=}5$ & $N{=}10$ & $N{=}1$ & $N{=}5$ & $N{=}10$ \\
    \hline
    ROMP~\cite{romp} & & 29.0 & \cellcolor{tabsecond}{32.1} & \cellcolor{tabsecond}{33.5} & \cellcolor{tabsecond}{34.8} & \cellcolor{tabsecond}{43.0} & \cellcolor{tabfirst}\bf{43.6} & \cellcolor{tabfirst}\bf{44.2} \\
    BEV~\cite{bev} & & 35.8 & \cellcolor{tabthird}{36.6} & \cellcolor{tabthird}{37.8} & 39.1 & 48.6 & \cellcolor{tabthird}{48.9} & \cellcolor{tabthird}{49.9} \\
    \hline
    Hand4Whole~\cite{hand4whole} & {\checkmark} & 77.9 & 73.3 & 366.5 & 733.0 & \cellcolor{tabfirst}\bf{26.3} & 98.3 & 188.3 \\
    OSX~\cite{osx} & {\checkmark} & 102.9 & 54.6 & 273.5 & 546.0 & 94.8 & 440.8 & 873.5 \\
    \hline
    \bf{\Ours-S} & {\checkmark} & 32.4 & \cellcolor{tabfirst}\bf{28.0} & \cellcolor{tabfirst}\bf{28.6} & \cellcolor{tabfirst}\bf{28.8} & \cellcolor{tabthird}44.4 & \cellcolor{tabsecond}{44.5} & \cellcolor{tabsecond}{44.6} \\
    \bf{\Ours-B} & {\checkmark} & 99.0 & 38.0 & 38.9 & \cellcolor{tabthird}{39.0} & 143.9 & 144.2 & 144.4 \\
    \bf{\Ours-L} & {\checkmark} & 318.7 & 50.8 & 50.9 & 50.9 & 478.7 & 479.5 & 479.8 \\
\end{tabular}\hspace*{-0.3cm}
    }
    }
    \end{minipage}
}
\end{table*}

%% file: sections/05_conclusion.tex
\section{Conclusion}
\label{sec:conclusion}

We presented \Ours, the first single-shot method for multi-person whole-body human mesh recovery.
It estimates accurate expressive 3D meshes (body, face and hands) and 3D positions in the scene, outperforming the state of the art for each sub-problem. 
Our model also adapts to camera information (\ie, intrinsics) when available.
\Ours is conceptually simple: it relies on a vanilla ViT backbone and a newly introduced cross-attention-based head for predictions.

%% file: sections/99_suppmat.tex
{\LARGE \noindent \textbf{Appendix} \vskip10pt}

This supplementary material contains additional implementation details and descriptions of the datasets and metrics used in the main paper (Appendix~\ref{sec:data_metrics}), details about how our synthetic \dataset dataset was generated (Appendix~\ref{sec:booster_data}), additional quantitative results (Appendix~\ref{sec:results}) and ablation studies (Appendix~\ref{sec:ablations_supp}), and finally, a discussion on limitations (Appendix~\ref{sec:limitations}). 
We also attached an additional video to showcase some results obtained with \Ours.

\section{Implementation, Datasets and Metrics}
\label{sec:data_metrics}
In this section, we give details about implementation, as well as each dataset used in the main paper, followed by a detailed description of the evaluation metrics.

\subsection{Implementation details}
\label{sec:implem_details}
By default, we use squared input images of resolution $448{\times}448$, with the longest side resized to $448$ and the smallest zero-padded to maintain aspect ratio. The only data augmentation used is random horizontal flipping.
The weights of the backbone are initialized with DINOv2~\cite{dinov2}.
We experiment with Small, Base and Large ViT models as encoder,
with a batch-size of 8 images and an initial learning rate of 5e-5. Our models are trained  with automated mixed precision~\cite{amp} for 400k iterations.
At resolution $448{\times}448$, training a ViT-S (resp. ViT-L) takes around 2 (resp. 5) days on a single NVIDIA V100 GPU.
The default detection threshold is $\tau{=}0.5$.
We use the neutral SMPL-X model~\cite{choutas2020expose} with $10$ shape components.

\subsection{Datasets descriptions}

\mydataset{BEDLAM~\cite{bedlam}} is a large-scale multi-person synthetic dataset composed of 300k images for training including diverse body shapes, skin tones, hair and clothing.
Synthetic humans are built by using a SMPL-X mesh and adding some assets such as clothes and hair. 
In each scene there are between 1 to 10 people with diverse camera viewpoints, and the test set is composed of 16k images.

\mydataset{AGORA~\cite{agora}} is a multi-person high realism synthetic dataset which contains 14k images for training, 2k images for validation and 3k for testing.
It consists of 4,240 high-quality human scans each fitted with accurate SMPL and SMPL-X annotations.
Results on the test set are obtained using an online leaderboard for SMPL and SMPL-X results.
We also report results on the validation for the distance estimation following~\cite{romp,bev} since the leaderboard does not give this metric on the test set.

\mydataset{3DPW~\cite{3dpw}} is an outdoor multi-person dataset composed of 60 sequences which contain respectively 17k images for training, 8k images for validation and 24k images for testing.
It was the first in-the-wild dataset in this domain for evaluating body mesh reconstruction methods~\cite{kolotouros2019spin,li2022cliff}.

\mydataset{MuPoTs~\cite{mupots}} is an outdoor multi-person dataset captured in a multi-view setting.
The dataset is composed of 8k frames from 20 real-world scenes with up to three subjects.
We use this dataset for evaluation only.
Poses are annotated in 3D with 14 body joints.

\mydataset{CMU Panoptic~\cite{cmu}} is a large-scale controlled environment multi-person dataset captured using multiple cameras.
Each person is annotated with 14 joints in 3D.
Following prior works~\cite{psvt,crmh}, we use 4 sequences which leads to a test set composed of 9k images.

\mydataset{EHF~\cite{ehf}} is the first evaluation dataset for SMPL-X based models.
It was built using a scanning system followed by a fitting of the SMPL-X mesh.
It is a single person whole-body pose dataset composed of 100 images.

\mydataset{UBody~\cite{osx}} is a large-scale dataset covering a wide range of real-life scenarios such as fitness videos, VLOGs or sign language.
Most of the time only the upper body part of the persons is visible.
We use the inter-scene protocol where there are 55k images for training and 2k images for testing.

\mydataset{Training datasets used by state-of-the-art methods} are many, and each method uses its own mix. For more transparency, we report in Table~\ref{tab:train_data} the training sets used by all methods that we compare to in Table 5 of the main paper.
\input{tab/sota_train_sets}

\subsection{Metrics Descriptions}

Prior work on multi-person human mesh recovery proposed metrics that can be separated into three categories: i) metrics that evaluate the reconstruction of the human mesh, centered around the root joint; ii) metrics that evaluate detection and iii) metrics that evaluate the prediction of spatial location.
In this section, we review the metrics used in the main paper.

\myparagraph{Human-centered mesh metrics}
To evaluate the predicted human mesh, we center both estimated and ground-truth human meshes around the pelvis joint.
We use per-vertex error (PVE) to evaluate the accuracy of the entire 3D mesh.
When available, we also report PVE computed on vertices corresponding to the face and hands only (PVE-Face and PVE-Hands). 
Because global orientation mistakes heavily impact the PVE, we also assess prediction quality without taking the global orientation into account by reporting all these metrics after Procrustes-Alignment (denoted with the prefix PA).
Since some human body datasets do not have mesh ground-truths but only 3D keypoints, we also report Mean Per Joint Position Error (MPJPE) on the 14 LSP 3D keypoints as well as the Percentage of Correct Keypoints (PCK) using a threshold of 15cm.

\myparagraph{Detection metrics}
To evaluate detection we rely on the Recall, Precision and F1-Score metrics.
On some datasets, it is also common to report  normalized mean joints error (NMJE) and normalized mean vertex error (NMVE), which are obtained by dividing mean joint errors and mean vertex errors by the F1-Score. This produces a score sensitive to both reconstruction quality and detection.

\myparagraph{Spatial location metrics}
To evaluate distance predictions we use the Mean Root Position Error (MRPE) by using the pelvis as root keypoint.

\subsection{Universal model and Fine-tuning strategy}
In the main paper, Table 5 presents the performance of a universal model (denoted with a $\dagger$) on multiple benchmarks, and results obtained by fine-tuning the model on a specific training set.
The universal model is trained on a combination of BEDLAM, AGORA, \dataset and UBody. The UBody dataset contains noisy ground truths, unlike  BEDLAM, AGORA and \dataset. Nevertheless we found that for the universal model, including UBody in the training data improves robustness to in-the-wild images with little impact on synthetic benchmarks.
This was not the case for other real-world datasets such as MS-CoCo or MPII, possibly because they have the same annotation issues but bring less variability.
For results reported with finetuning, we follow the standard practice of independently finetuning on the training set of AGORA, 3DPW and UBody when evaluating on the respective benchmarks.
While CMU and MuPoTs do not have an associated training set, we still consider a simple finetuning strategy:
we finetune the universal model on BEDLAM, AGORA and 3DPW, by sampling images equally between the datasets during the finetuning stage. 
We observe that this brings substantial gains, presumably because this training data mix is better aligned with the data distributions of CMU and MuPoTs.

\input{sections/043_synthetic_data}

\section{Additional results}
\label{sec:results}

We now present additional results on two additional test datasets namely BEDLAM~\cite{bedlam} and 3DMPB~\cite{huang2022pose2uv}.

\myparagraph{Results on BEDLAM-test}
We report results on BEDLAM-test in Table \ref{tab:bedlam} using the recently released online leaderboard.
Since the leaderboard is extremely recent (online since October 2023), we were unable to compare to many existing methods.
At the time of this submission,
only single-person methods~\cite{pixie,li2022cliff} are reported in the leaderboard which makes the comparison with our method difficult.
Still, \Ours significantly outperforms other methods on this datasets.

\begin{table*}[t]
    \centering
    \caption{\textbf{BEDLAM-test leaderboard}. CLIFF is trained on BEDLAM, CLIFF+ on BEDLAM+AGORA. \emph{\Ours is the only multi-person and the only single-shot method reported on the benchmark to date.}}
    \vspace{-0.2cm}
    \resizebox{\linewidth}{!}{
    \begin{tabular}{l|ccc|ccccc}
         Method&  F1-Score$\uparrow$ & Precision$\uparrow$ & Recall$\uparrow$ & Body-MVE$\downarrow$ & FullBody-MVE$\downarrow$ & Face-MVE$\downarrow$ & LHand-MVE$\downarrow$ & RHand-MVE$\downarrow$  \\
         \hline
         PIXIE~\cite{pixie} & 0.94 & \bf{0.99} & 0.90 & 100.8 & 149.2 & 51.4 & 44.8 & 48.9 \\
         CLIFF~\cite{li2022cliff}& 0.94 & \bf{0.99} & 0.90 & 61.3 & 94.6  & 29.8 & 34.7 & 35.5 \\
         CLIFF+~\cite{li2022cliff}& 0.94 & \bf{0.99} & 0.90& 57.5 & 87.2 & 27.3 & 30.3 & 32.6 \\
         \bf{\Ours} & \bf{0.97} & \bf{0.99} & \bf{0.90} & \bf{53.4} & \bf{76.8} & \bf{21.3} & \bf{23.0} & \bf{25.8} \\
    \end{tabular}
    }
    \label{tab:bedlam}
\end{table*}

\myparagraph{Results on 3DMPB}
We report results on whole-body predictions on the 3DMPB dataset~\cite{huang2022pose2uv} in Table~\ref{tab:3dmpb}.
\Ours reaches state-of-the-art performance  with all backbones (ViT-S/B/L). 

\begin{table}[t]
\centering
\caption{\textbf{
Comparison to the state of the art on the 3DMPB dataset.
}}
\vspace{-0.2cm}
\begin{tabular}{l|c}
Method & PA-MPJPE$\downarrow$ \\
 \hline
ROMP~\cite{romp} & 72.0 \\
Pose2UV~\cite{huang2022pose2uv} & 69.5 \\
\bf{\Ours} ViT-S & {62.6} \\
\bf{\Ours} ViT-B & {55.8} \\
\bf{\Ours} ViT-L &  {\bf{49.7}} \\
\end{tabular}
% \vspace{-0.3cm}
\label{tab:3dmpb}
\end{table}

\section{Additional ablations}
We conduct additional ablations on model design choices. First, we evaluate various initializations for ViT-Base models. Second, we ablate different choices of primary keypoint in our formulation of detection. Third, we evaluate the impact of the different training losses considered in the main paper.
\label{sec:ablations_supp}
\subsection{Backbone pretraining}
In Figure~\ref{fig:pretraining}, we report results using various pretraining methods, with a ViT-Base architecture and $448{\times448}$ input images. 
DINO~\cite{dinov1} and DINOv2~\cite{dinov2} rely on self-supervised pre-training, while ViT-Pose~\cite{vitpose} is trained with 2D body keypoints supervision.
DINOv2 leads to the best final performance, and converges faster. 
The difference in performance decreases while training longer, which may be due to the relatively large size of our training set, with ViT-Pose eventually achieving comparable results. Thus, using DINOv2 may be most beneficial when training compute is limited.

\begin{figure}[b!]
\centering
\begin{tikzpicture}[baseline=0cm]
\begin{axis}[
    width=0.7\linewidth,
    height=3.5cm,
    ylabel= {\normalsize MuPoTs \\ PCK3D ($\uparrow$)},
    ylabel shift = -0.2cm,
    xlabel = Iterations,
    xlabel shift = -0.1cm,
    xmin=70000, xmax=830000,
    ylabel style={align=center, font=\normalsize},
    xlabel style={align=center, font=\normalsize},
    tick label style={font=\normalsize},  
    xticklabels = {0, 0, 100k,, 300k,,, 600k,, 800k},
    every axis plot/.append style={ultra thick},
    legend pos=south east,
    x tick label style={yshift={0.05cm}},
    scaled x ticks = false,
    legend style = {column sep=2pt, font=\scriptsize, only marks, opacity=1},
]
    \addplot[style=dinov2] coordinates {
            (100000, 54.3)
            (300000, 57.4)
            (600000, 70.1)
            (800000, 73.6)
            };
    \addlegendentry{DINO}
    \addplot[style=vitpose] coordinates {
            (100000, 58.0)
            (300000, 65.8)
            (600000, 72.3)
            (800000, 74.1)
            };
            \addlegendentry{ViT-Pose}
               \addplot[style=dino,opacity=1] coordinates {
            (100000, 69.9)
            (300000, 72.9)
            (600000, 75.8)
            (800000, 75.9)
            };
            \addlegendentry{DINOv2}
    
\end{axis}
\end{tikzpicture}
\vspace*{-0.3cm}
\caption{\textbf{Impact of backbone pretraining}.
Initializing the backbone with DINOv2 leads to faster convergence. 
% \vspace{0.2cm}
}
\label{fig:pretraining}
\end{figure}

\input{tab/91_ablations_suppmat}

\subsection{Choice of primary keypoint} In Table~\ref{tab:primary_kpt}, we report results with different choices of primary keypoint: 
\emph{head},  \emph{pelvis} or \emph{spine}.
The method appears robust to this choice, though using the head as primary keypoint yields better results by a small margin.
We postulate it might be due to the fact that the head is less often occluded in images, and we keep the head as primary keypoint.

\subsection{Training losses on 3D and 2D} We experiment with different combinations of reconstruction losses: directly on the SMPL-X parameters (\textit{rot}), on the vertices produced by the SMPL-X model (\textit{v3d}), a combination of both (\textit{rot + v3d}), and the addition of reprojection losses (+\textit{v2d}). 
Table~\ref{tab:loss} shows that adding as much supervision as possible (in 3D, 2D and rotation space) yields the best performance, possibly because it reduces ambiguities during training.

\section{Limitations}
\label{sec:limitations}

While \Ours reaches state-of-the-art performance across multiple human mesh recovery benchmarks, we still observe some limitations that may be improved upon in the future.

\myparagraph{Patch-level detection}
We follow the CenterNet~\cite{zhou2019objects} paradigm for the detection stage, which allows us to propose a single-shot method without elaborate post-processing.
However it comes with the main limitation that multiple humans (i.e. person-centers) may belong to the same patch in the image. Because of this, some collisions happen during training and some detections are impossible  at inference time.
This well-known limitation is already discussed in Appendix C of the CenterNet paper~\cite{zhou2019objects}.
We refer reader to this section for more details.
In our case we observe that as long as we use images of reasonable resolution (i.e. more than $448{\times}448$) and a small patch-size (\ie, $14{\times}14$), collisions remain very rare at training.
As shown in the attached video, \Ours produces reasonable predictions even in relatively crowded environment which indicates that our modeling is overall robust.
In future work, robustness could likely be increased further, \eg by having multiple queries per patch.

\myparagraph{Truncated humans}
We observe that \Ours sometimes struggles to detect humans when the head is not visible; this may in part be due to the fact that we chose the head as primary keypoint, and also to the fact that such data is very rare in the training datasets.
We still observe (see attached video) that \Ours is able to detect human when only a small part of the head is visible.
We make the assumption that adding more aggressive cropping augmentations during training would lead to a model more robust to this type of truncation by the image frame. 
As shown in the attached video, we observe that \Ours is already quite robust to occlusions in general.

\myparagraph{SMPL-X representation}
We employ the SMPL-X parametric 3D model for representing whole-body human mesh.
 As discussed in the method section, we use the pose parameters expressed by the relative rotations of each joint regarding its parent given a pre-defined kinematic tree.
Such representation is easy to use and commonly relied upon in practice~\cite{kolotouros2019spin,goel2023humans4d}, however it may raise several concerns: i) in general rotations are not easy to regress as they lie in a non-Euclidean space~\cite{bregier2021deepregression}.
This is a topic that may not have been explored sufficiently in the 3D vision community so far and may deserve further work -- we use the 6D representation for regression --
ii) regressing the pose using a relative representation can lead to an accumulation of errors, particularly on the extreme parts of the body (hands, legs).
We believe that investigating different pose representations would be beneficial for \Ours and the human mesh recovery field in general.

\begin{figure*}
    \centering
    \includegraphics[width=\linewidth]{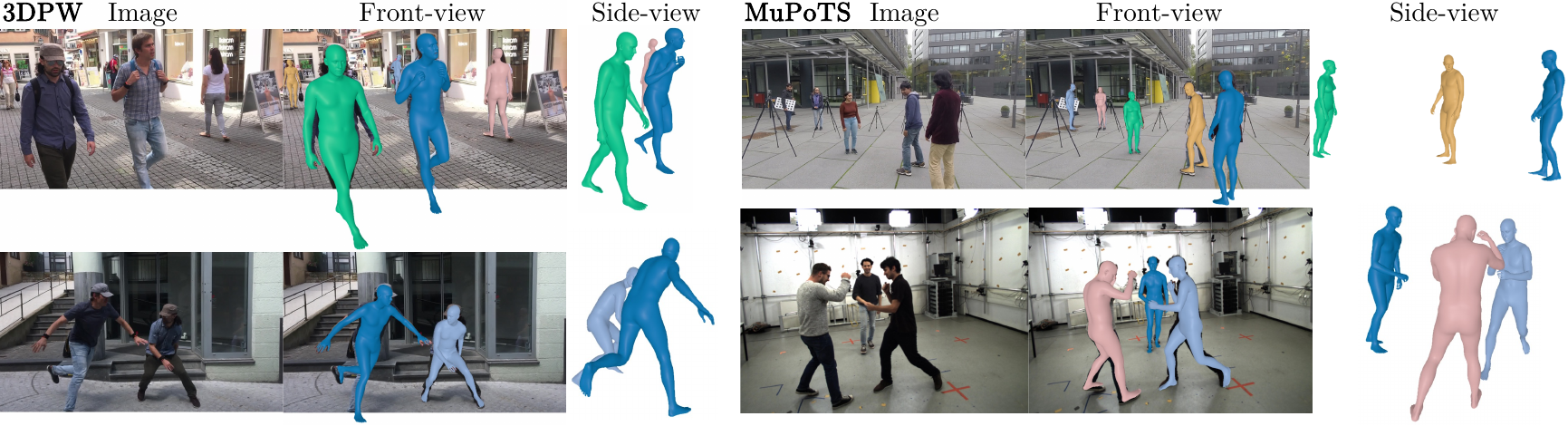} \\[-0.2cm]
    \caption{\textbf{Additional qualitative results of \Ours.}
    Front-view and Side-view 3D reconstructions on test images (\textit{Left}: 3DPW, \textit{Right}: MuPoTs).}
    \label{fig:side-view}
\end{figure*}

%% file: tab/sota_train_sets.tex
\begin{table*}[thb]
\centering
\hspace{-1em}
\centering
\caption{\textbf{Training datasets used by state-of-the-art models}.
ROMP~\cite{romp} mentions other datasets for training their `advanced' model, that we did not include. We also did not include hands-only or face-only datasets.
} \vspace{-0.2cm}
\resizebox{\linewidth}{!}{
\begin{tabular}{l@{\hspace{-0.1em}}cccccccccccc}
\bf{Method} & Human3.6M & MPI-INF-3DHP & PoseTrack & LSP & LSP Extended & MPII & MS-CoCo & MuCo-3DHP & CrowdPose & UP & AICH & RH  \\

\shline
\rowcolor{gray!10} \textit{Body-only} &&&&&&&&&&&& \\
CRMH~\cite{crmh}   & \cmark & \cmark & \cmark & \cmark & \cmark & \cmark & \cmark &&&&&    \\
3DCrowdNet~\cite{3dcrowdnet}  & \cmark &&&&& \cmark & \cmark & \cmark & \cmark &&&\\
ROMP~\cite{romp}  & \cmark & \cmark && \cmark && \cmark & \cmark &&&\cmark & \cmark & \\
BEV~\cite{bev} & \cmark &&& \cmark && \cmark & \cmark & \cmark & \cmark &&& \cmark  \\
PSVT~\cite{psvt} & \cmark &&& \cmark && \cmark & \cmark & \cmark & \cmark &&& \\
\shline
\rowcolor{gray!10} \textit{Whole-Body} &&&&&&&&&&&& \\
Hand4Whole~\cite{hand4whole} & \cmark &&&&& \cmark & \cmark &&&&&  \\
OSX~\cite{osx}  & \cmark &&&&& \cmark & \cmark &&&&& \\
SMPLer-X~\cite{cai2023smplerx}  & \multicolumn{11}{c}{\bf{32 datasets. Refer to their paper for a full list}}  \\
ExPose~\cite{choutas2020expose} & \cmark &&& \cmark & \cmark & \cmark &&&&&&  \\
FrankMocap~\cite{rong2021frankmocap} & \cmark &&&&&& \cmark &&&&&  \\
PIXIE~\cite{pixie} &&&& \cmark & \cmark & \cmark & \cmark &&&&&  \\
PyMAF-X~\cite{pymafx2023} & \cmark & \cmark && \cmark & \cmark & \cmark & \cmark &&&&&  \\
\end{tabular}
}
\label{tab:train_data}    
\end{table*}

%% file: sections/043_synthetic_data.tex
\begin{figure}[t]
\centering
\includegraphics[width=0.32\linewidth]{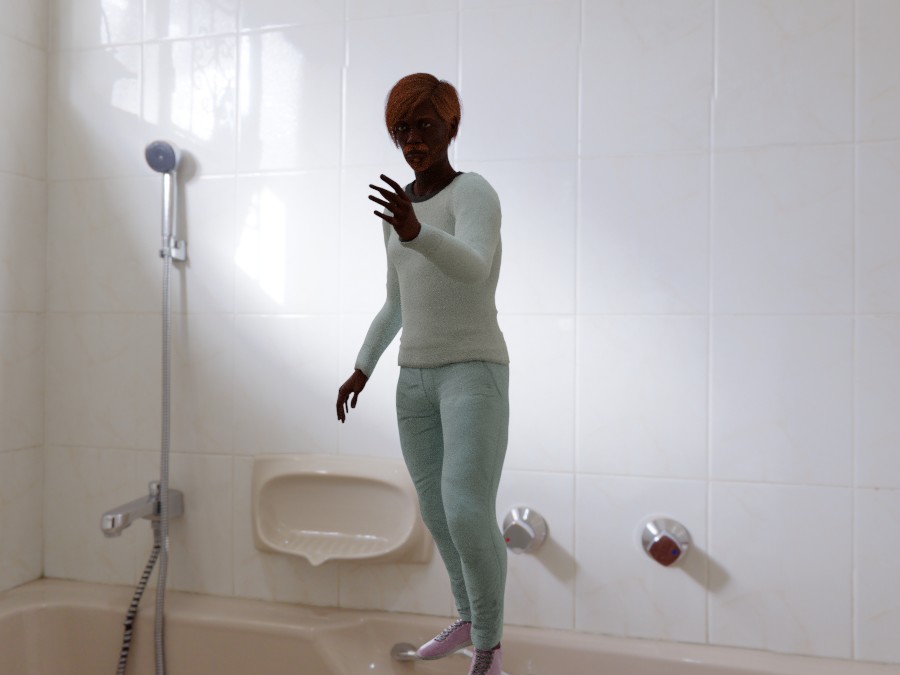}  \hfill
\includegraphics[width=0.32\linewidth]{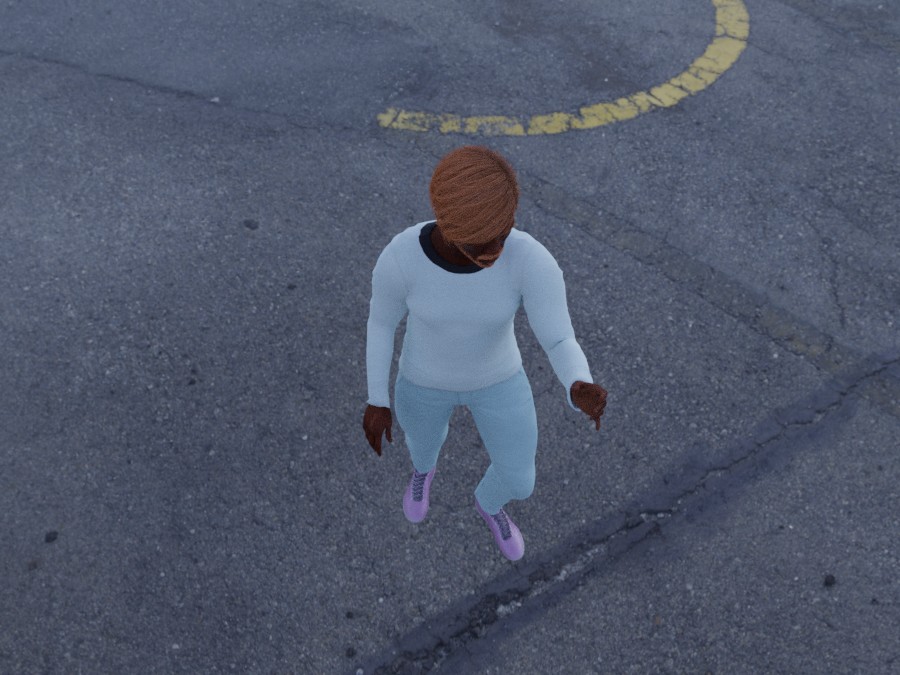}  \hfill
\includegraphics[width=0.32\linewidth]{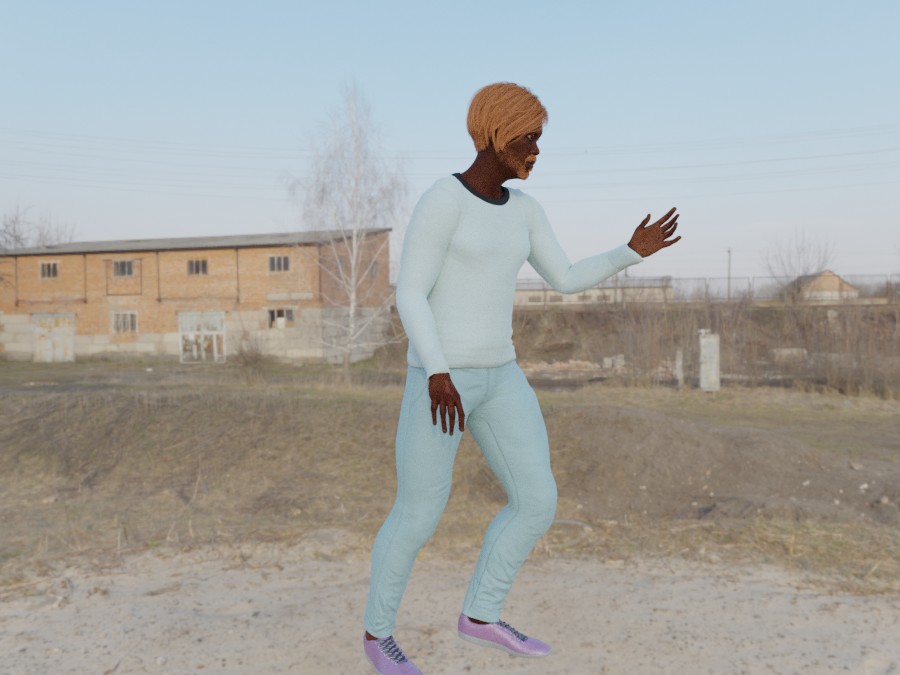} \\
\includegraphics[width=0.32\linewidth]{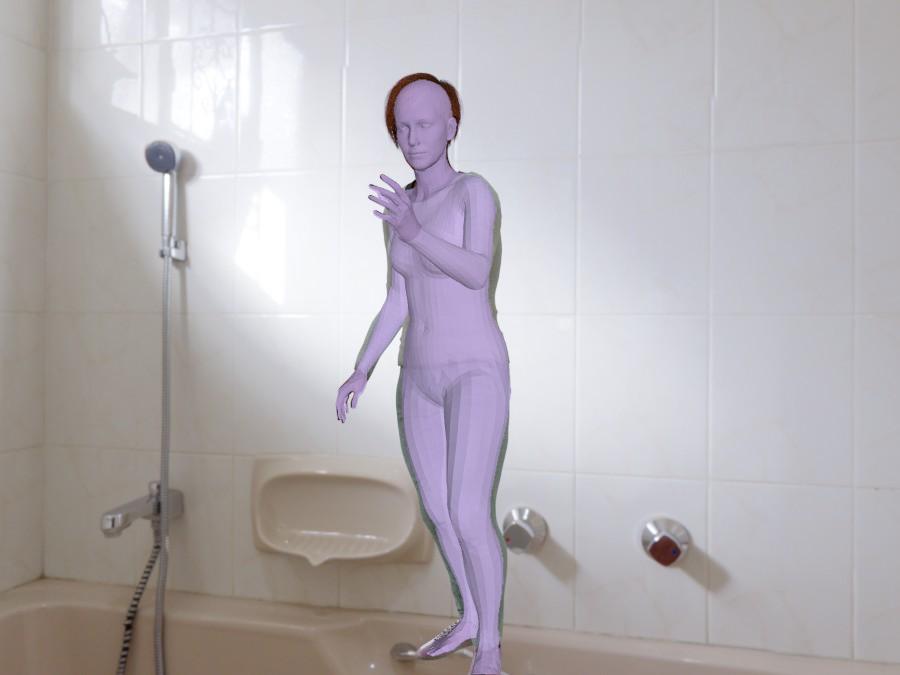}  \hfill
\includegraphics[width=0.32\linewidth]{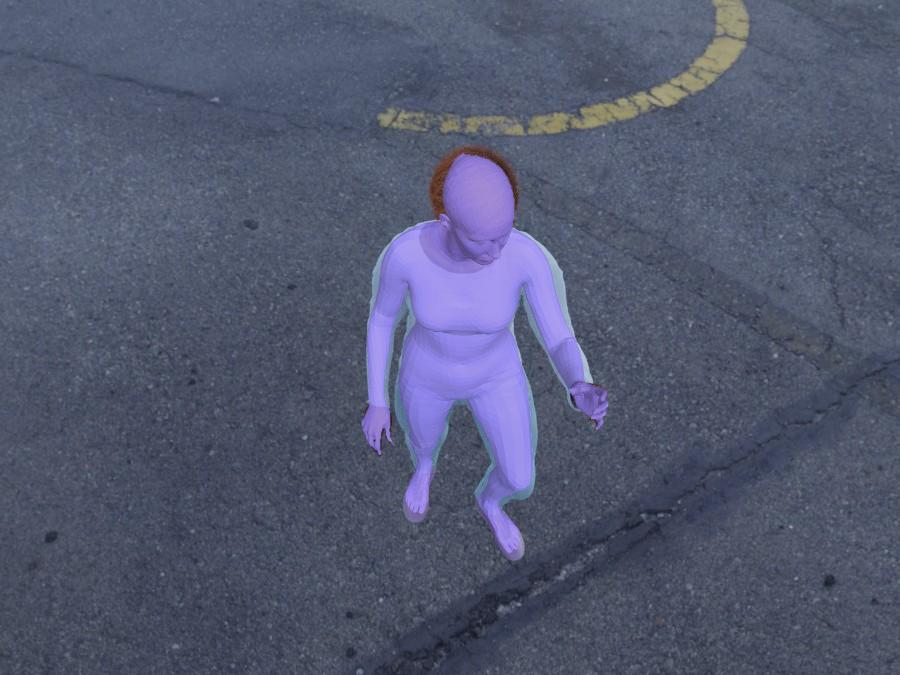}  \hfill
\includegraphics[width=0.32\linewidth]{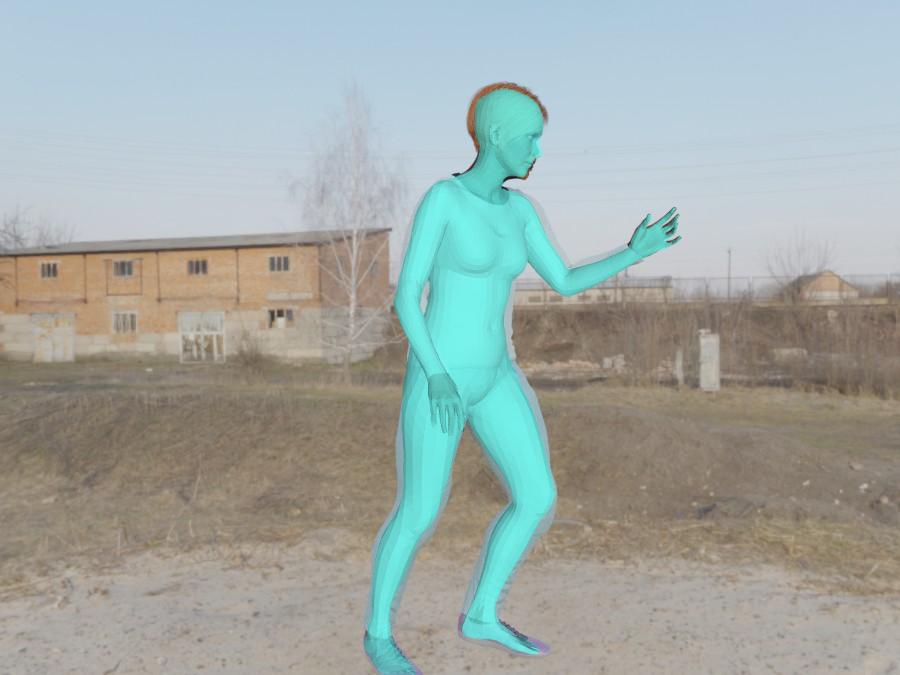} \\[-0.2cm]
\caption{
\textbf{Samples from our \dataset dataset} with a rendered human using HumGen3D (top) and the corresponding SMPL-X shape used for retargeting (overlaid at the bottom).
}
\label{fig:retargetting}
\end{figure}

\begin{figure*}[t]
\resizebox{\linewidth}{!}{
\begin{tabular}{@{}c@{ }c@{ }c@{ }c@{}}
Image & Overlaid SMPL-X GT & Close-up on hand (image) & Close-up on hand (GT) \\
\includegraphics[width=0.24\linewidth]{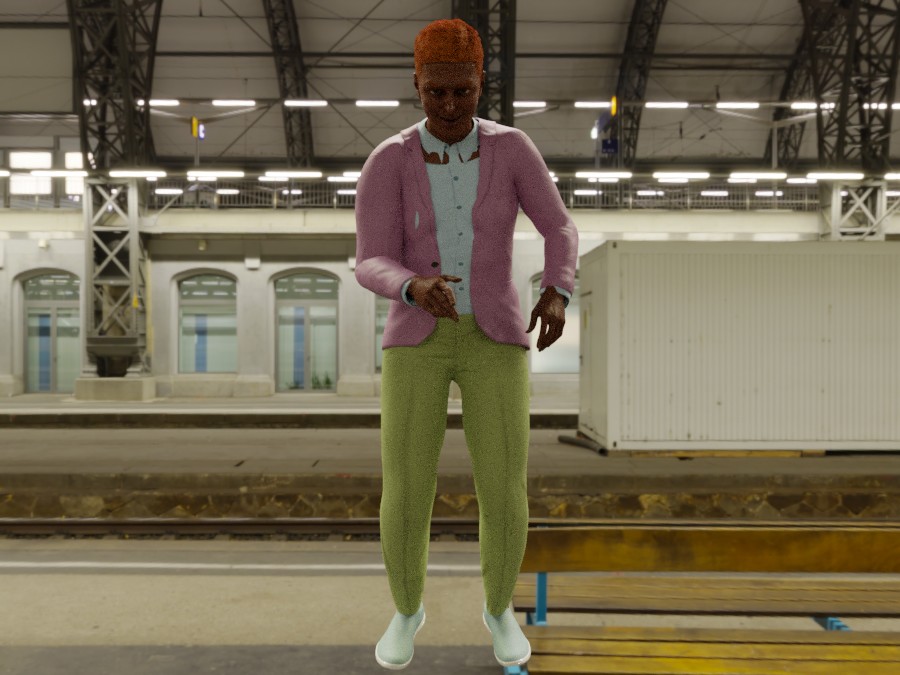} &
\includegraphics[width=0.24\linewidth]{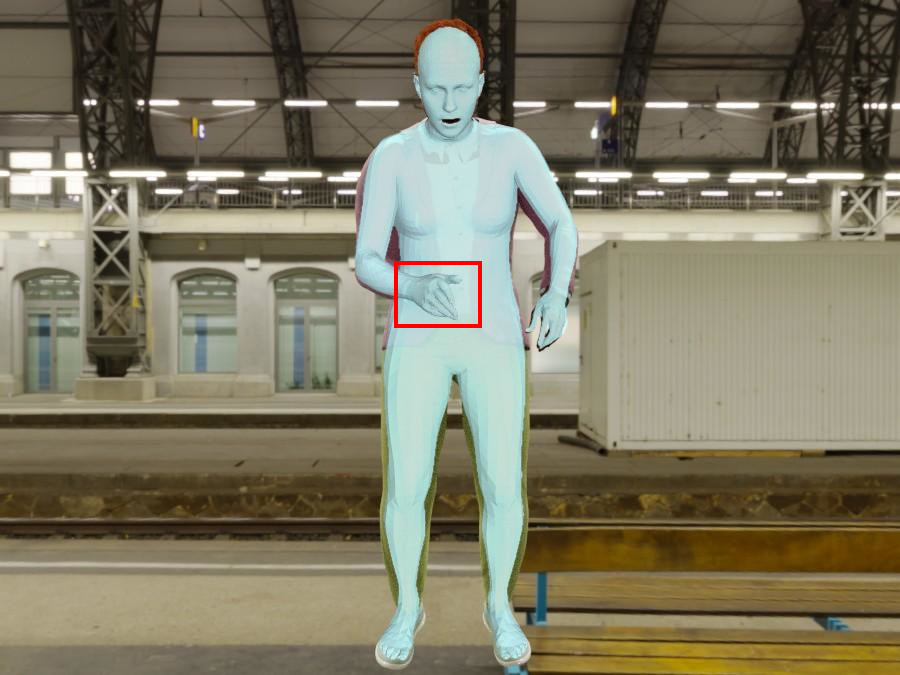} &
\includegraphics[width=0.24\linewidth]{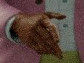} &
\includegraphics[width=0.24\linewidth]{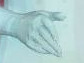} \\
\includegraphics[width=0.24\linewidth]{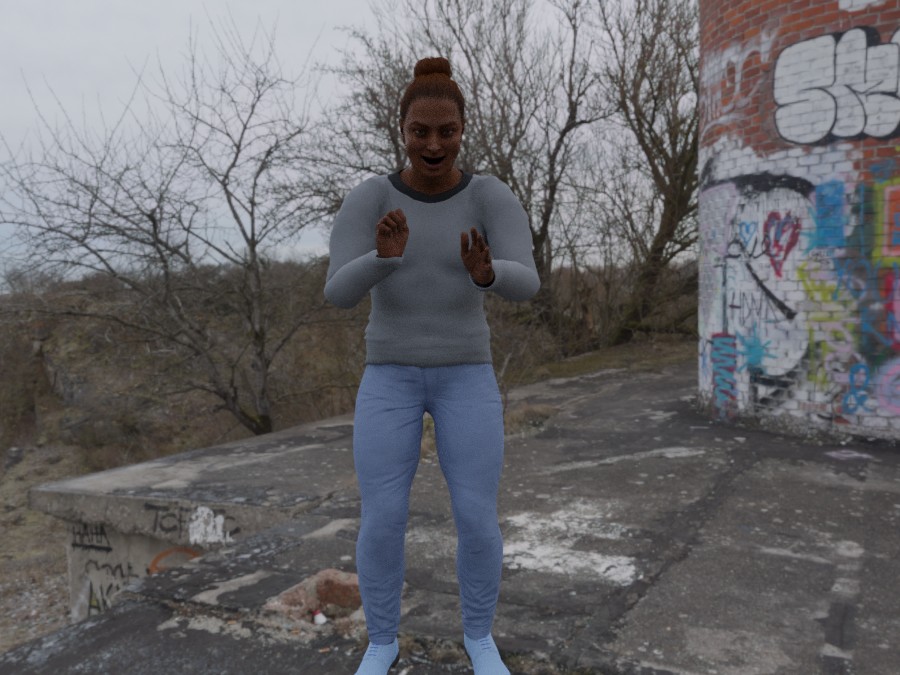} &
\includegraphics[width=0.24\linewidth]{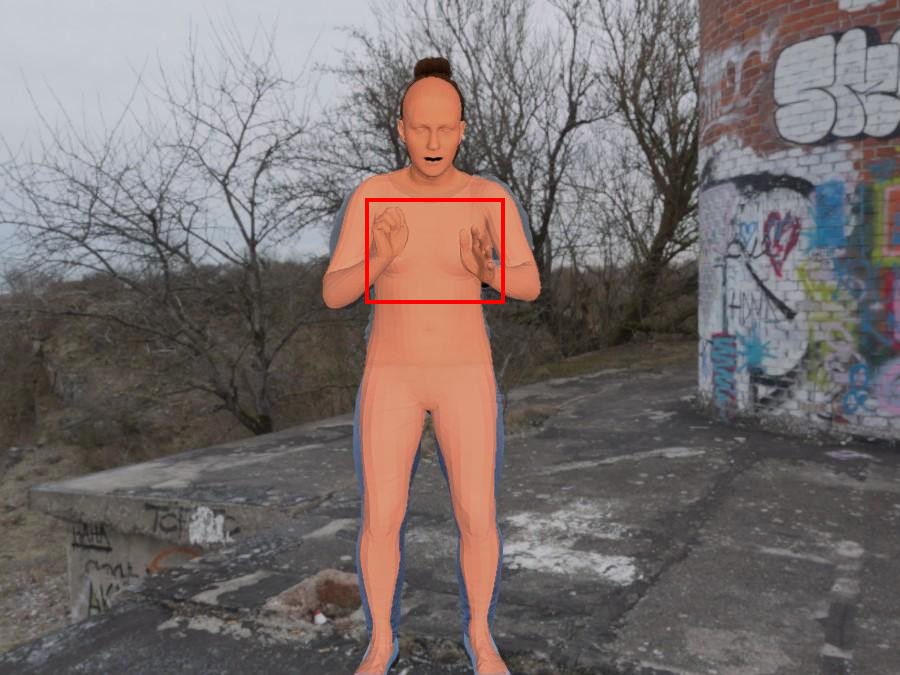} &
\includegraphics[width=0.24\linewidth]{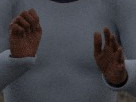} &
\includegraphics[width=0.24\linewidth]{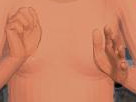} \\
\includegraphics[width=0.24\linewidth]{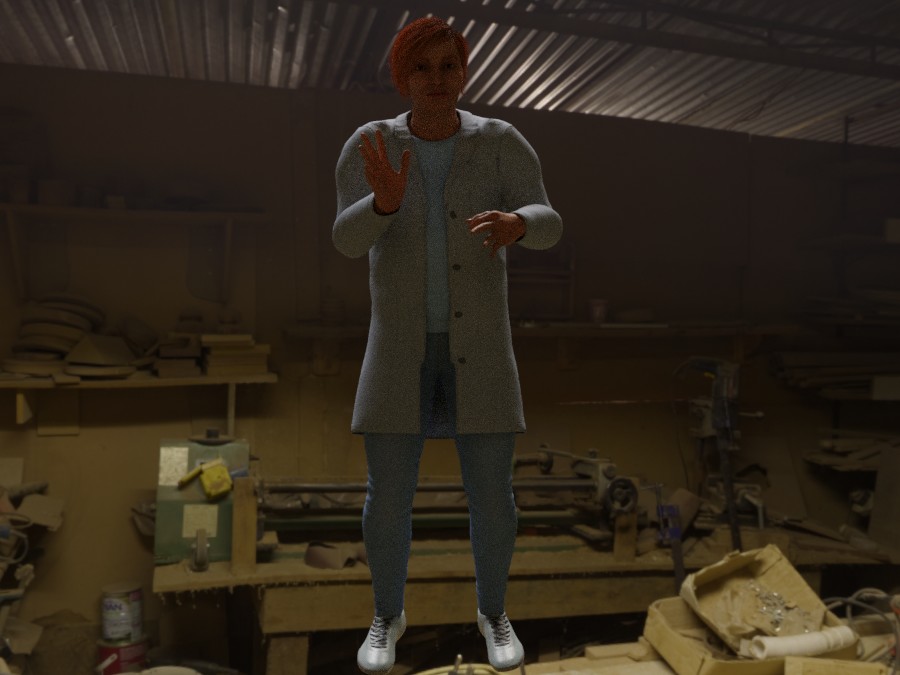} &
\includegraphics[width=0.24\linewidth]{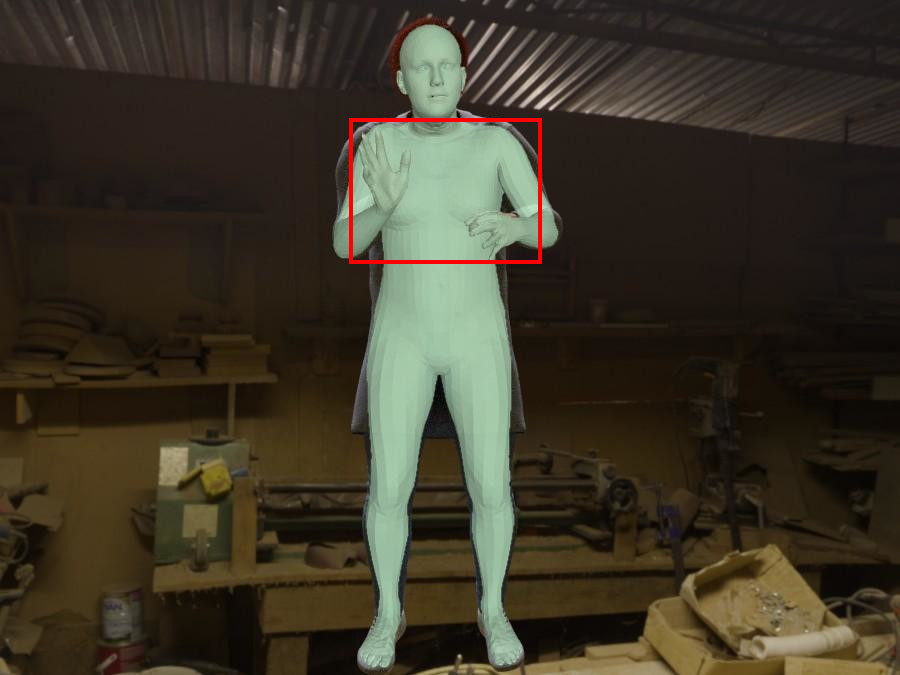} &
\includegraphics[width=0.24\linewidth]{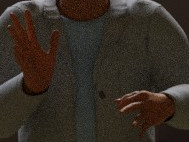} &
\includegraphics[width=0.24\linewidth]{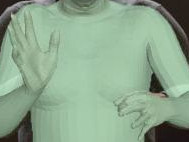} \\[-0.4cm]
\end{tabular}
}
\caption{\textbf{Examples from the \dataset dataset} showing the input image, the overlaid SMPL-X annotations, a close-up on the image and annotations around the hands corresponding to the rectangle shown in the second column. People are seen up close, and diverse hand poses are used.}
\label{fig:hands_examples}
\end{figure*}

\begin{figure*}[t]
\centering
\resizebox{\linewidth}{!}{
\begin{tabular}{@{}c@{ }c@{ }c@{ }c@{ }c@{ }c@{}}
\includegraphics[width=0.13\linewidth]{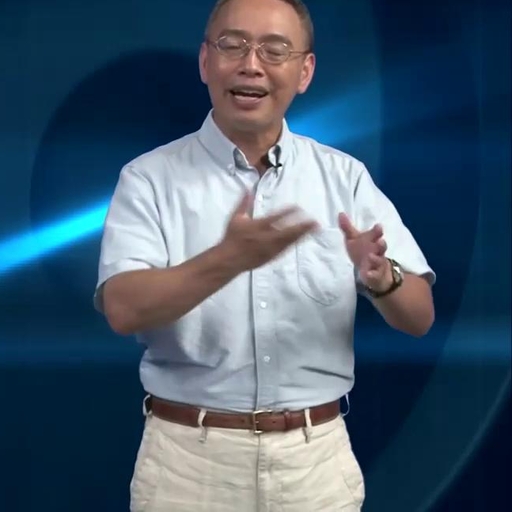}  &
\includegraphics[width=0.13\linewidth]{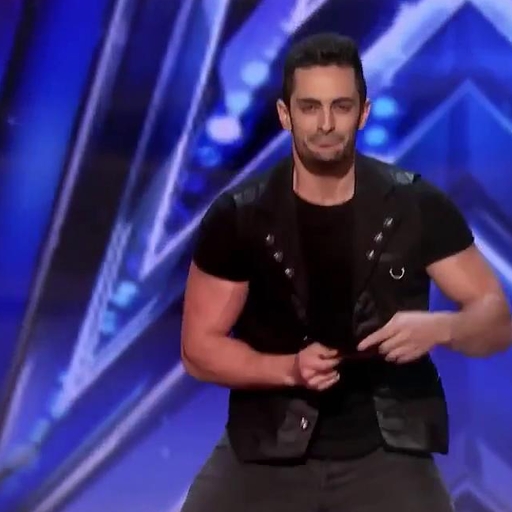} &
\includegraphics[width=0.13\linewidth]{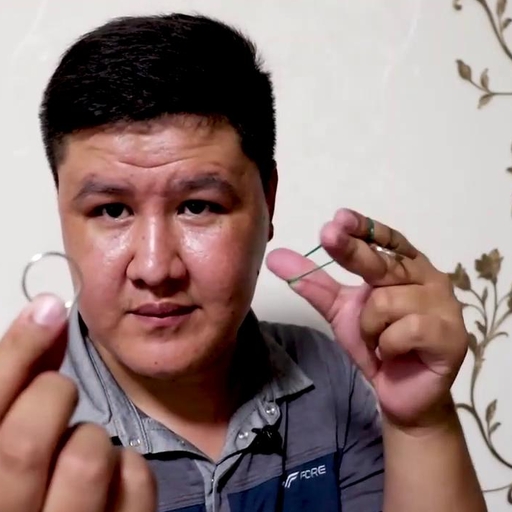} &
\includegraphics[width=0.13\linewidth]{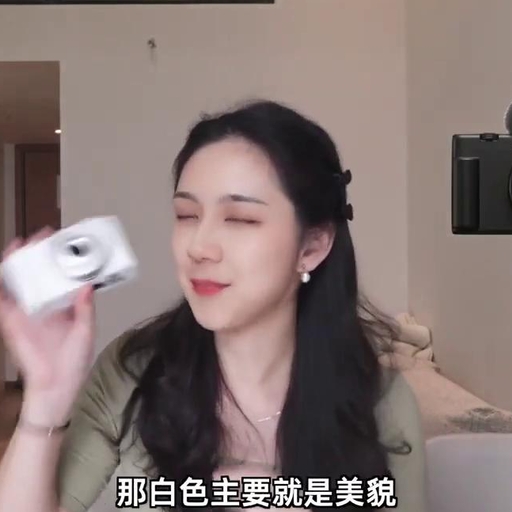} &
\includegraphics[width=0.13\linewidth]{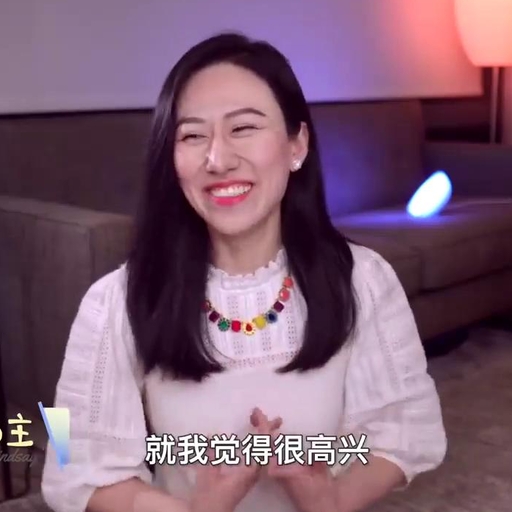} &
\includegraphics[width=0.13\linewidth]{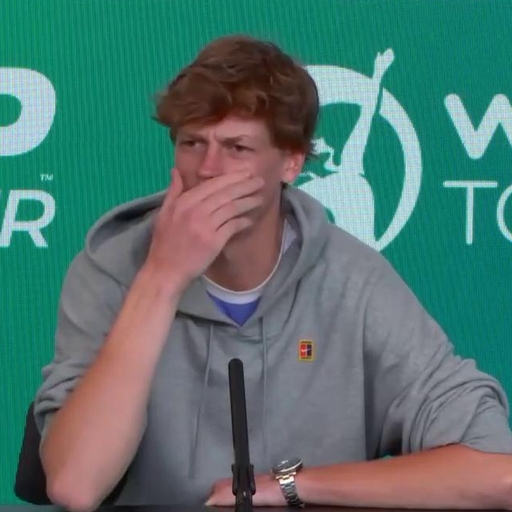} \\

\frame{\includegraphics[width=0.13\linewidth]{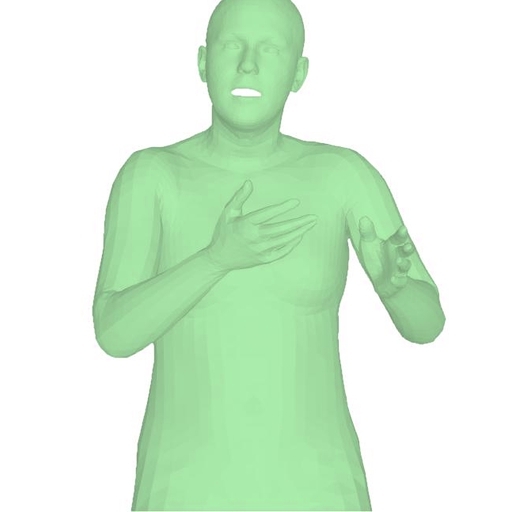}} &
\frame{\includegraphics[width=0.13\linewidth]{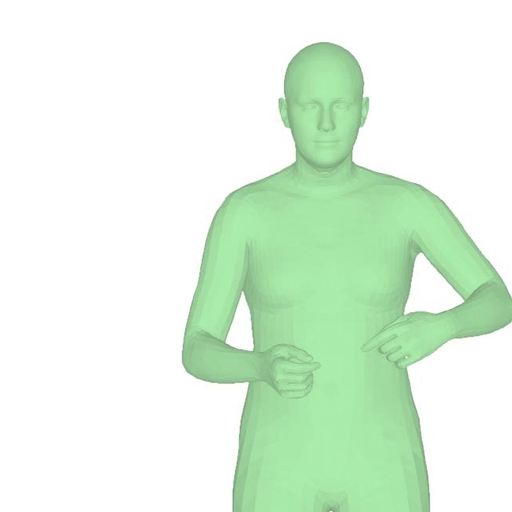}} &
\frame{\includegraphics[width=0.13\linewidth]{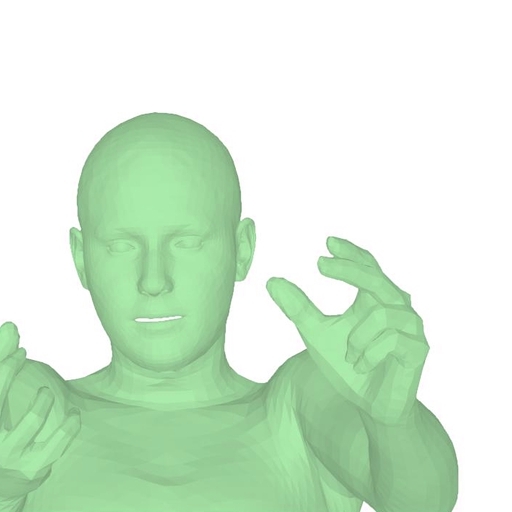}} &
\frame{\includegraphics[width=0.13\linewidth]{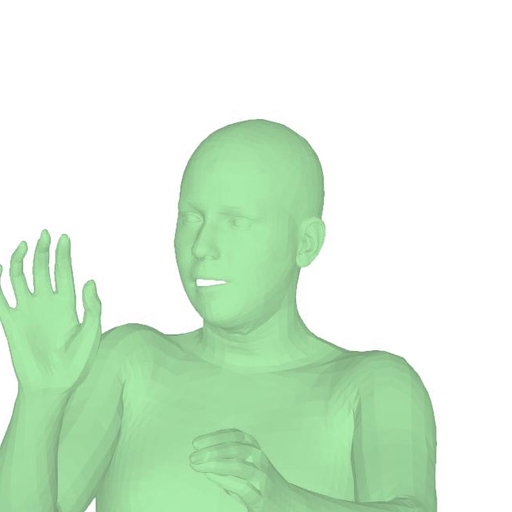}} &
\frame{\includegraphics[width=0.13\linewidth]{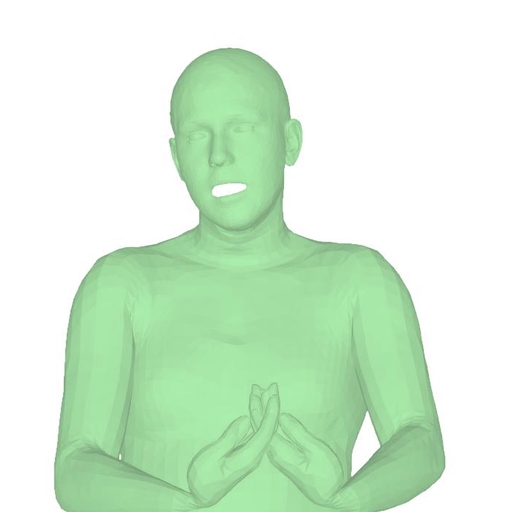}} &
\frame{\includegraphics[width=0.13\linewidth]{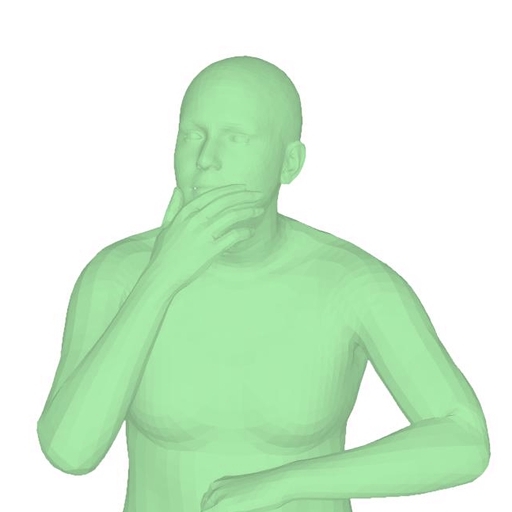}} \\

\frame{\includegraphics[width=0.13\linewidth]{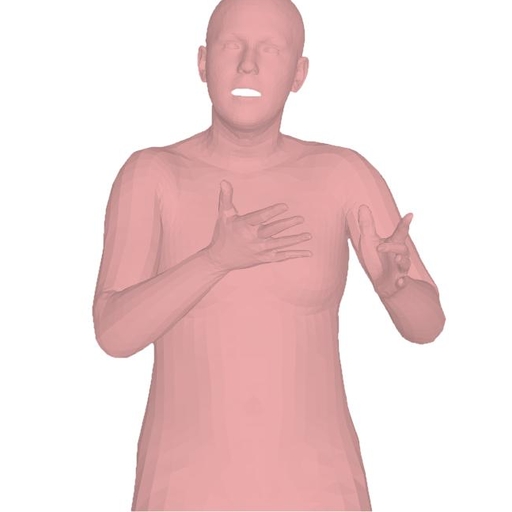}} &
\frame{\includegraphics[width=0.13\linewidth]{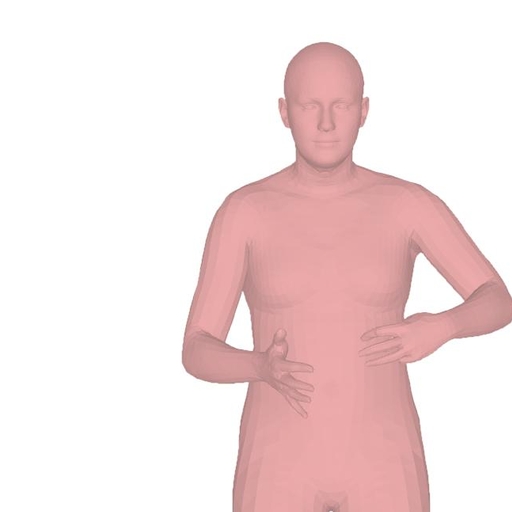}} &
\frame{\includegraphics[width=0.13\linewidth]{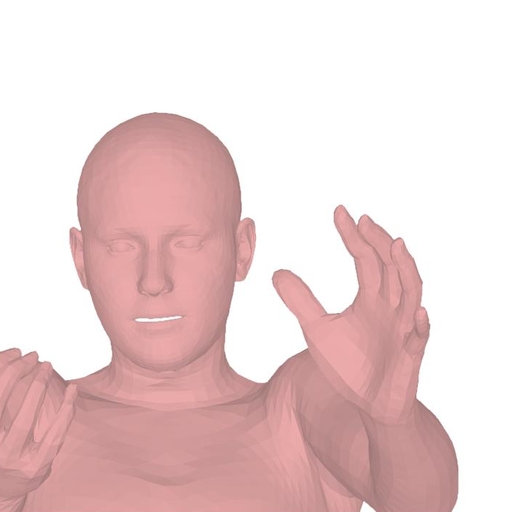}} &
\frame{\includegraphics[width=0.13\linewidth]{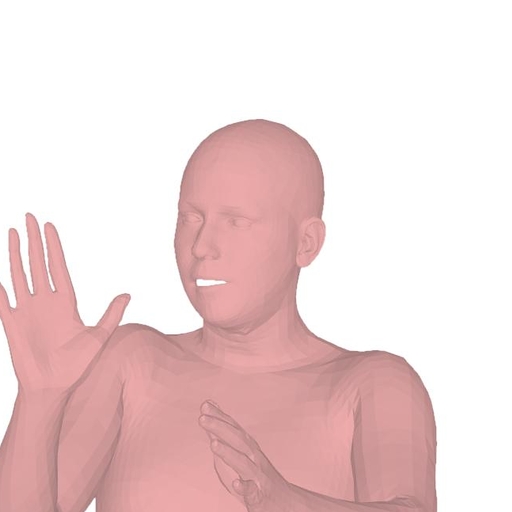}} &
\frame{\includegraphics[width=0.13\linewidth]{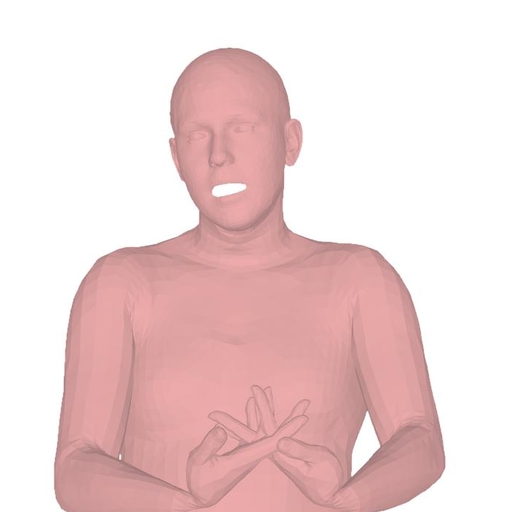}} &
\frame{\includegraphics[width=0.13\linewidth]{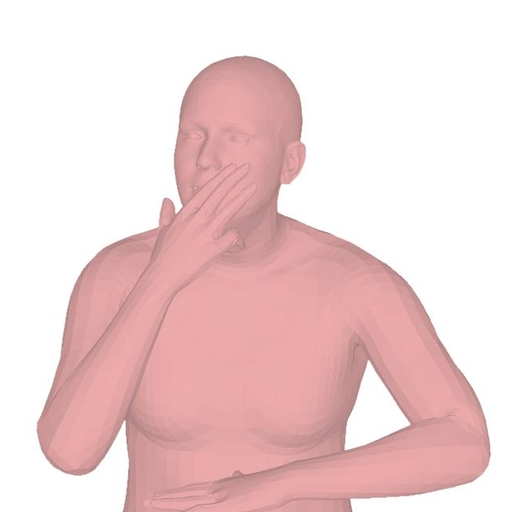}}
\end{tabular}
}
\vspace{-0.4cm}
\caption{
\textbf{Illustrations of how we increase hand diversity in human shape sources to be rendered.} Given an annotation from UBody (image on top, annotation in the middle row), we swap the hands from a large set built from InterHand2.6M to have more diversity in terms of hand poses.} 

\label{fig:hands_inpainting_example}
\end{figure*}

\section{The synthetic \dataset dataset}
\label{sec:booster_data}

\myparagraph{Motivation}
Existing synthetic datasets, namely BEDLAM and AGORA, provide perfect ground truths for the SMPL-X model, \ie, including faces and hands. 
However, in these datasets: i) most humans are seen from afar, which is not ideal to capture subtle details needed to properly reconstruct faces and hands and ii) hand poses lack diversity.
In particular since our method is single-shot, \ie, runs without specific image crops or feature resampling around hands, hands consist of only a few visible pixels for many training images.
We remedy this by adding a dedicated, booster dataset, consisting of close-up pictures of single humans with clearly visible hands in diverse poses, to the rest of the training data.

\myparagraph{3D Human models}
We render images of 3D human models.
Following the strategy of BEDLAM~\cite{bedlam}, we use a procedural generation pipeline with fine control over parameters, rather than commercially available scans of clothed humans (\eg as in AGORA~\cite{agora}).
To this end, we make use of HumGen3D~\cite{humgen3d}, 
a human generator add-on to the Blender software tool~\cite{blender}.
This add-on generates 3D rigged human models, with different clothing (layered on top of the body mesh), hairstyles, skin tones, age, \etc This yields a high diversity of humans overall.

\myparagraph{SMPL-X annotations}
In order to produce precisely annotated images, we take SMPL-X parameters as input and deform human models to closely match these annotations. We proceed through iterative optimization by minimizing the pairwise distance between corresponding points at the surface of SMPL-X and human mesh models, using semi-automatically annotated dense correspondences.
Figure~\ref{fig:retargetting} shows examples of rendered avatars and their associated SMPL-X meshes and illustrates the quality of the annotations.

\myparagraph{Rendering}
Characters are placed in 
empty scenes 
with random 
high dynamic range images 
from Poly Haven~\cite{polyhaven} as environment 
backgrounds.
We render images with a $900{\times}675$ resolution and a 56.2\textdegree{} horizontal field of view.
The principal point is set at the center of the image.

\myparagraph{Human shape sources and hand diversity}
We seek to generate humans that are: i) close to the camera such that the hands are sufficiently visible, and ii) with diverse hand poses.
For the first point, we simply render images of a single person, facing the camera, at a distance varying slightly around $2.5$ meters so that it fills most of the image.
We find that this yields clearly visible hands. 
For the second point, we sample human poses from BEDLAM, AGORA, and UBody, where hand annotations are respectively: taken from the GRAB~\cite{grab} dataset, fitted to 3D scans, and fitted to in-the-wild images. 
In addition to these three sources, in order to further diversify our set of hand poses, we also augment UBody's annotations with hands from other sources: we create a large set of diverse hand poses using MANO~\cite{mano} annotations from the InterHand2.6M \cite{interhand} dataset. This is done by extracting all MANO annotations and converting them into a right hand format, using a mirroring operation for left hand poses. When creating a synthetic image with augmented hands, we sample two random 
hand annotations from the large set, transform one into a left hand format and replace 
SMPL-X hand annotations using the new hand poses.
This left/right augmentation strategy further increases hand pose diversity compared to the original InterHand2.6M dataset.

\myparagraph{Dataset}
We generate about $60$k images, 
with human shapes equally sampled from i) BEDLAM, ii) AGORA, iii) UBody, iv) UBody, with increased hand diversity.
We show qualitative examples of our generated images and the associated SMPL-X mesh in Figures~\ref{fig:retargetting} and~\ref{fig:hands_examples}. We also provide examples of hand 
pose augmentations
in Figure~\ref{fig:hands_inpainting_example}.

\begin{figure*}[t]
\resizebox{\linewidth}{!}{
\begin{tabular}{c@{ }c@{ }c@{ }c} 
Input image & Zoom & w/ \dataset & w/o \dataset \\
\includegraphics[width=0.24\linewidth]{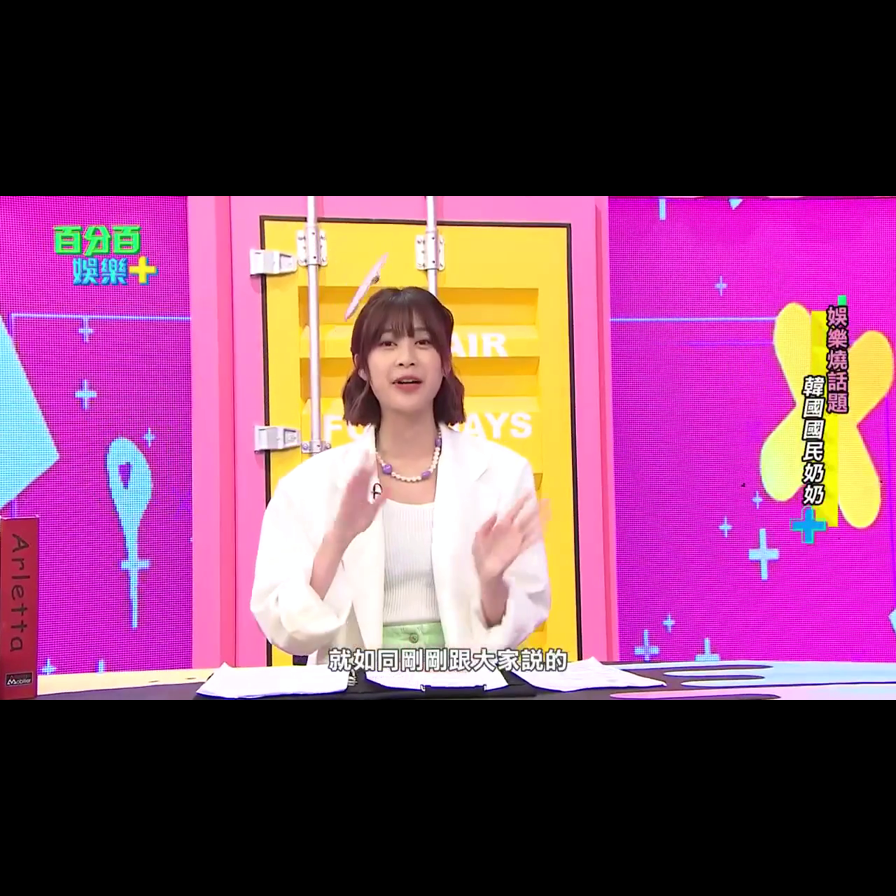} &
\includegraphics[width=0.24\linewidth]{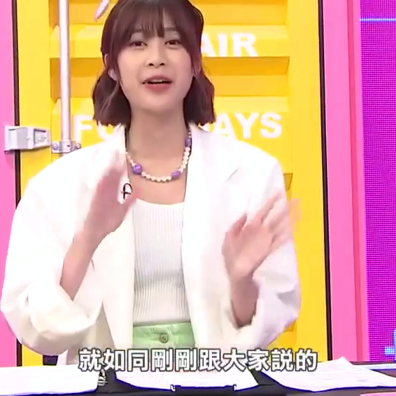} &
\frame{\includegraphics[width=0.24\linewidth]{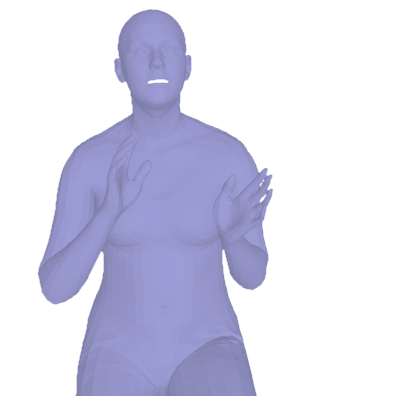}} &
\frame{\includegraphics[width=0.24\linewidth]{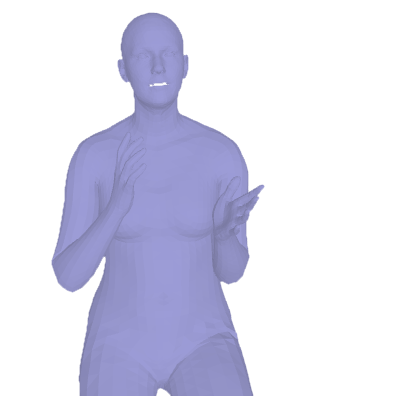}} \\
\includegraphics[width=0.24\linewidth]{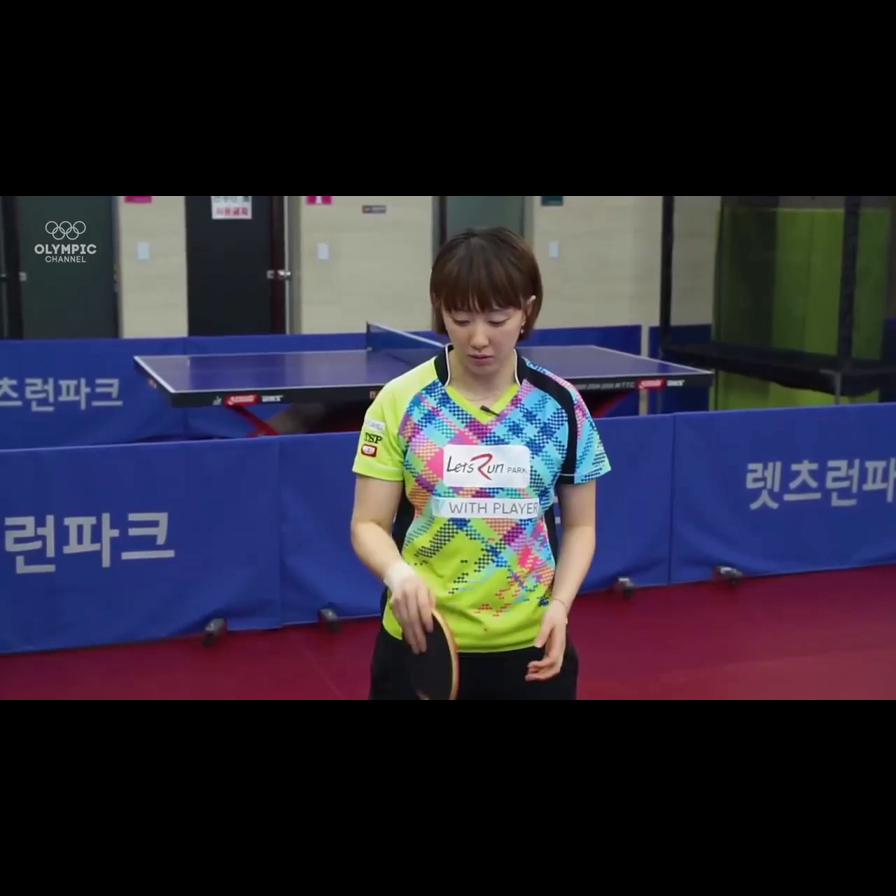} &
\includegraphics[width=0.24\linewidth]{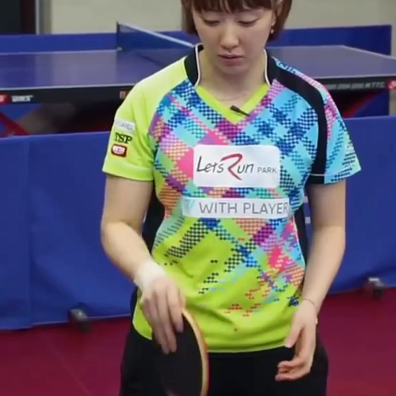} &
\frame{\includegraphics[width=0.24\linewidth]{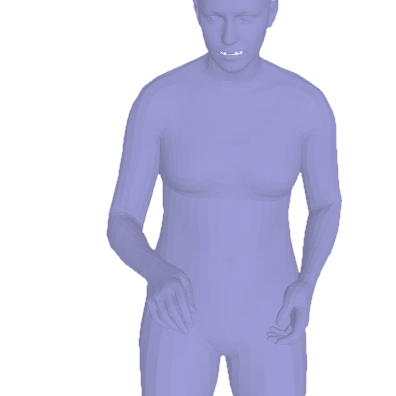}} &
\frame{\includegraphics[width=0.24\linewidth]{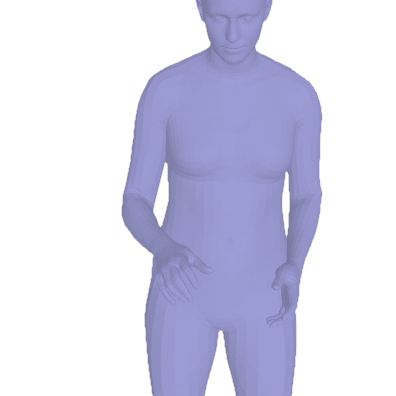}} \\
\includegraphics[width=0.24\linewidth]{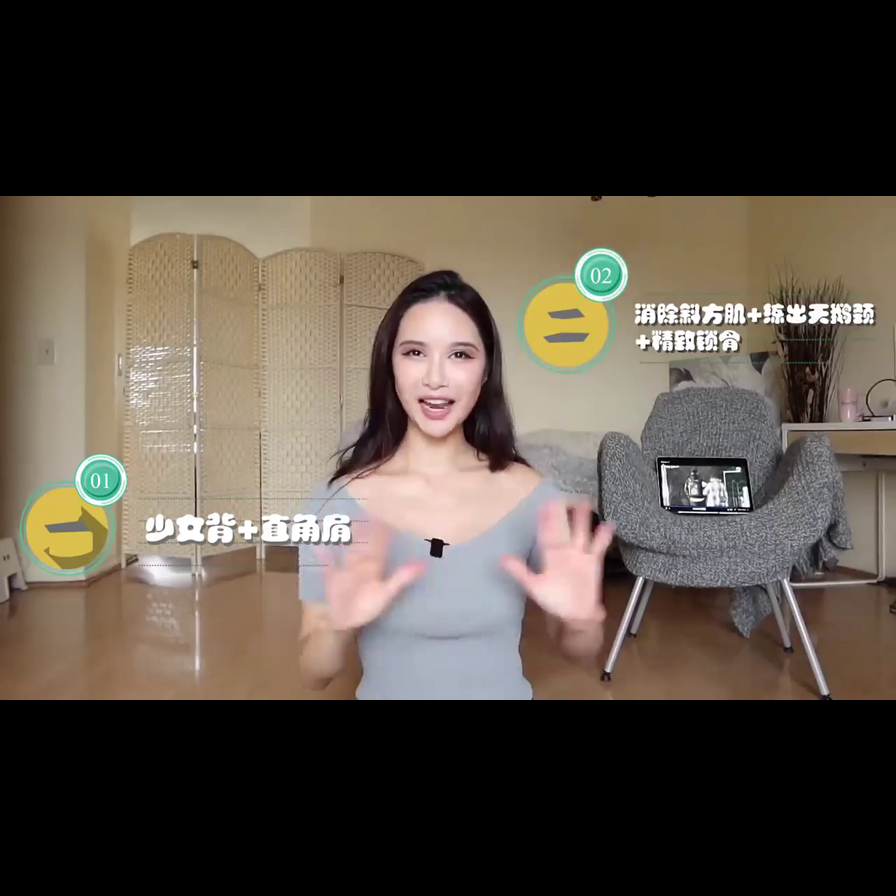} &
\includegraphics[width=0.24\linewidth]{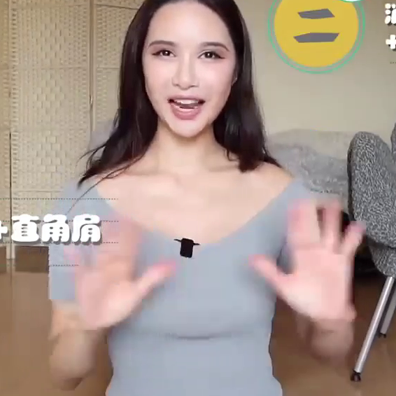} &
\frame{\includegraphics[width=0.24\linewidth]{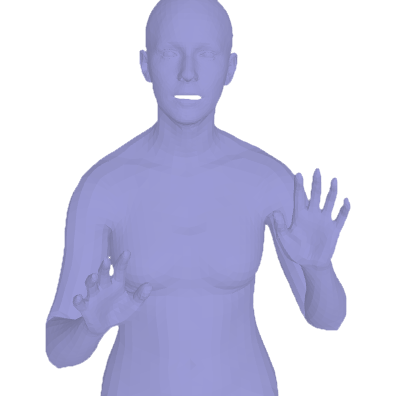}} &
\frame{\includegraphics[width=0.24\linewidth]{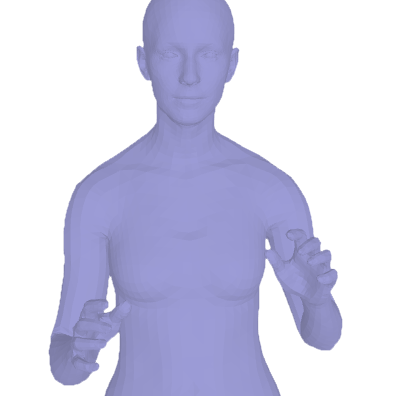}} \\[-0.3cm]
\end{tabular}
}
\caption{\textbf{Qualitative results on some UBody images with or without training with our synthetic \dataset dataset}. Hand pose predictions are more accurate when the model has been trained with the synthetic \dataset dataset.}
\label{fig:quali_hands}
\end{figure*}

\myparagraph{Impact} 
In addition to the quantitative gain reported in the main paper, we show some qualitative example of adding the synthetic \dataset dataset to the training set in Figure~\ref{fig:quali_hands}.
For instance, in the third example, the hands are significantly better predicted when the training set includes our synthetic \dataset dataset.

%% file: tab/91_ablations_suppmat.tex
\begin{table*}[t]
\centering
\caption{
\textbf{Additional ablative study}.
{
We report additional ablation experiments on \textbf{(a)} the choice of the primary keypoint, and \textbf{(b)} the influence of training losses.
}
\vspace{-0.2cm}
}
\hspace*{-0.1cm}
\subfloat[
\textbf{Primary keypoint}
\label{tab:primary_kpt}
]{
\begin{minipage}{0.45\linewidth}{
\begin{center}
\tablestyle{2pt}{1.05}
\renewcommand{\arraystretch}{1.137}
\begin{tabular}{y{22}x{34}x{24}x{24}}
 & {\scriptsize MuPoTS$\uparrow$} & {\scriptsize 3DPW$\downarrow$} & {\scriptsize EHF$\downarrow$}\\
\shline
\baseline{head}  & \baseline{76.3} & \baseline{\bf{74.4}} & \baseline{\bf{55.3}}\\
pelvis & 77.0 & 74.5 & 57.5 \\
spine1 & \bf{77.1} & 74.9 & 56.1 \\
spine3 & 76.5 & 74.8 & 56.8 \\
\end{tabular}
\end{center}
}
\end{minipage}
}
\hfill
\subfloat[
\textbf{Losses}
\label{tab:loss}
]{
\begin{minipage}{0.45\linewidth}{
\begin{center}
\tablestyle{2pt}{1.05}
\renewcommand{\arraystretch}{1.137}
\begin{tabular}{y{28}x{34}x{24}x{24}}
 & {\scriptsize MuPoTS$\uparrow$} & {\scriptsize 3DPW$\downarrow$} & {\scriptsize EHF$\downarrow$}\\
\shline
v3d &  75.0 & 76.1 & 65.0\\
rot & 70.1 & 92.2 & 97.9 \\
\baseline{~~+v3d} & \baseline{76.3} & \baseline{73.5} & \baseline{55.3}\\
~~+v2d & \bf{79.2} & \bf{70.5} & \bf{53.2}\\
\end{tabular}
\end{center}
}
\end{minipage}
}
\label{tab:ablations}
\end{table*}